\documentclass[pdflatex,sn-basic]{sn-jnl}

\usepackage{graphicx}%
\usepackage{multirow}%
\usepackage{amsmath,amssymb,amsfonts}%
\usepackage{amsthm}%
\usepackage{mathrsfs}%
\usepackage[title]{appendix}%
\usepackage{xcolor}%
\usepackage{textcomp}%
\usepackage{manyfoot}%
\usepackage{booktabs}%
\usepackage{algorithm}%
\usepackage{algorithmicx}%
\usepackage{algpseudocode}%
\usepackage{listings}%
\usepackage[section]{placeins}

\usepackage{algorithmicx}%
\usepackage{algpseudocode}%
\usepackage{listings}%
\usepackage{adjustbox}
\usepackage{subcaption}
\newcommand{\com}[1]{\textcolor{black}{#1}}

\begin{document}

\title[AL under Class Imbalance and Label Shift]{Active Learning under Simulated Class Imbalance and Label Shift in Binary Semantic Segmentation}

\author*[1,2]{\fnm{Julien} \sur{Combes}}\email{julien.combes@michelin.com}

\author[1]{\fnm{Jean-Francois} \sur{Coeurjolly}}\email{jean-francois.coeurjolly@univ-grenoble-alpes.fr}

\author[2]{\fnm{Alexandre} \sur{Derville}}\email{alexandre.derville@michelin.com}

\affil*[1]{\orgdiv{} \orgname{Laboratoire Jean Kuntzmann}, \orgaddress{\city{Grenoble}, \postcode{38000}, \country{France}}}

\affil[2]{\orgdiv{} \orgname{Manufacture Française des Pneumatiques Michelin}, \orgaddress{\street{Pl. des  Carmes Dechaux}, \city{Clermont-Ferrand}, \postcode{63000}, \country{France}}}


\abstract{

The aim of Active Learning is to select the most informative samples from an unlabelled set of data. This is useful in cases where the amount of data is large and labelling is expensive, such as in machine vision or medical imaging. Two particularities of machine vision are first, that most of the images produced are free of defects, and second, that the amount of images produced is so big that we cannot store all acquired images. This results, on the one hand, in a strong class imbalance in defect distribution and, on the other hand, in a potential label shift caused by limited storage. To understand how these two forms of  imbalance  affect active learning algorithms, we propose a simulation study based on two open-source datasets. \com{Our contribution is not a new acquisition strategy but a controlled and reproducible protocol that synthetically constructs datasets in which the levels of class imbalance and label shift are fixed and vary independently}. Four standard active learning selection strategies are compared: random sampling, entropy-based selection, core-set selection and an imbalance aware strategy called galaxy sampling. We empirically find that active learning, and in particular the entropy-based and core-set selections, remains competitive with and typically superior to random sampling even for highly imbalanced datasets, and we quantify the loss of efficiency that occurs under a strong label shift.

}

    
\maketitle
    
\section{Introduction}\label{sec1}
    
In industrial contexts, the collection of substantial datasets to address specific tasks is typically feasible. However, in the context of semantic segmentation for defect detection in quality control, most images are defect free due to the high reliability of production processes. For instance, an automatic visual inspection system (such as described in \cite{mignotAutomaticInspectionSystem2024}) placed at the end of a production line is exposed to mostly healthy images. Indeed, manufacturing processes are subject to rigorous quality control policies with the aim of minimising the number of defects along production lines. The process yields a substantial volume of data, yet only a limited portion is characterised by substantial informational content (in our case, this is evidenced by the presence of segmentation masks). This results in a marked class imbalance, wherein the defect class is noticeably underrepresented in comparison to the background or non-defective regions. This imbalance poses a significant challenge for learning approaches and is particularly critical in active learning settings, where the selection of informative samples might be biased towards the majority class, see e.g.~\cite{popDeepEnsembleBayesian2018}.\\
    
\com{\noindent{\bf Contributions.} This paper is a controlled sensitivity study rather than a methodological contribution. Specifically, we propose a reproducible protocol that synthesises binary segmentation datasets with independently controlled class imbalance (the faulty-image proportion in the unlabelled pool) and label shift (a mismatch between the faulty-image proportions in the pool and in the test set). We continue by benchmarking four standard acquisition strategies (random, entropy, core-set, galaxy) across a grid of imbalance and shift levels on three contrasting open-source datasets. Finally, we quantify, with significance testing, how the benefit of active learning erodes as the pool/test mismatch grows.\\
}
{The rest of the paper is organized as follows. Section~\ref{sec-relatedwork} reviews related work on active learning, class imbalance and label shift. Section~\ref{sec-method} presents the setting of pool-based active learning applied to semantic segmentation including main notation used in this manuscript. We also detail the active learning algorithms we consider, provide some implementation details and summarize the criteria we use to compare them. Section~\ref{sec-results} is the core of our paper. We propose a simulation study based on three open-source datasets. We specify how we construct, in a fair and reproducible way, datasets for which we control the class imbalance and label shift. Then, we apply the different AL algorithms to these generated datasets and present the results, before a discussion (Section~\ref{sec-discussion}) and a conclusion (Section~\ref{sec-conclusion}) providing recommendations and open questions in the context of machine vision}.
        
    \section{Related Work}\label{sec-relatedwork}
    
    \noindent{\bf Active learning.} Although substantial quantities of industrial data can be collected, a significant impediment pertains to the necessity for expert annotation, a process that is both costly and time-consuming, yet indispensable for the training of segmentation models. Active learning is proposed as a solution to reduce annotation effort by identifying the most informative samples, thereby allowing accurate models to be trained with minimal labelled data. Despite the fact that pioneering contributions to active learning (AL for short) in image processing emerged over a decade ago~\cite{joshiMulticlassActiveLearning2009}, there has been a considerable surge in interest in this domain with the advent of sophisticated deep segmentation architectures  \cite{maBreakingBarrierSelective2024a}. In the context of defect detection, it is noteworthy that the majority of images are found to be free of defects. It is important to note that these samples are frequently excluded from training due to the fact that segmentation networks are designed to operate solely on defective samples. In this paper, the investigation focuses on the influence of the proportion of defect-free images present in the unlabelled dataset on the effectiveness of active learning strategies for binary semantic segmentation. Two main frameworks coexist within active learning~\cite{renSurveyDeepActive2021}.
    The first is the \textbf{pool based} AL where we consider two sets of data, a set where the data is labeled and another non-annotated one. 
    In this setting, we consider a specific budget of data (number of unlabelled data that can be labeled)  that is queried from the unlabelled pool of data. A statistical model is trained on the newly updated dataset, iterating this process for a predefined number of cycles or until the labeling capacity is exhausted. The second approach is the \textbf{stream based} \cite{lughoferOnlineActiveLearning2017} AL where the data to be labeled is selected at the data acquisition time.

    The drawback of the pool-based setting is the long-term storage requirement for unlabelled datasets, which can quickly grow in size, particularly in machine vision applications. However, this setting ensures to get the most relevant and informative data to train the model. A balance must be struck between minimizing data storage and preserving potentially informative datasets. In the domain of machine vision applied to quality control, it is imperative to avoid the loss of informative images  during the training of deep learning models. This paper, therefore, focuses only on the sensitivity of pool-based AL algorithms. \\
    
    \noindent{\bf Class imbalance.} The class imbalance in computer vision is dependent on the problem trying to be solved. 
    Indeed, in image classification, each image corresponds to a certain class, so the weights are depending on the number of each image belonging to each class. 
    However, in semantic segmentation, the classification is done at the pixel level, so we should know what  the number of pixels belonging to each class is, which makes it more likely to lead to imbalance. 
    Many approaches exist to tackle the class imbalance problem in the parameters estimation phase such as~\cite{budaSystematicStudyClass2018}, like oversampling low frequency classes or undersampling frequent classes. In deep-learning the preferred methods rely on modifying the loss function, by weighting it depending on each class frequency, also known as cost sensitive learning~\cite{elkanFoundationsCostsensitiveLearning2001}. In the field of image segmentation, novel loss functions have been developed to address the challenges posed by difficult-to-classify pixels. A notable example is the focal loss proposed by~\cite{linFocalLossDense2018} which requires prior knowledge of the class distribution and/or tuning hyperparameters. In the context of AL, where the availability of labelled data is limited, it is imperative to retain as much data as possible for the training process.
    The descriptive analysis of Active Learning on imbalanced data algorithms remains rare. Numerous studies concentrate on AL algorithms that operate on imbalanced datasets~\cite{aggarwalActiveLearningImbalanced2020,zhangGALAXYGraphbasedActive2022,zhaoActiveLearningLabel2021} to our knowledge, only one applied those concepts on semantic segmentation~\cite{yuanLearningHardtolearnActive2024}.
    Some research has been conducted into running simulation studies, and we could find one involving synthetic class imbalance in image classification by removing images from certain categories on the CIFARs datasets~\cite{choiVaBALIncorporatingClass2020} but none using real life machine vision data.  However, it is difficult to quantify imbalance in these settings when comparing different levels of imbalance. Indeed, we don't know whether the introduced imbalance is the source of the observed results or whether the same results would have been obtained with other modifications to the distributions. Creating synthetic imbalance in the multiclass setting is challenging. \\
    
    \noindent{\bf Label Shift.} In industry, most of produced images are healthy and only faulty ones are used for training (i.e. 100\% of images in the training set contain defects). Two factors contribute to this circumstance. Primarily, storage capacities are constrained, requiring a preliminary  selection of images to be uploaded and subsequently saved for future labeling. Secondly, in the industrial sector, despite the rarity of defects, their identification is of paramount importance. Industries cannot afford leaks that could lead to adverse consequences. Training on primarily faulty data can skew the distribution towards false positive while minimizing leaks. However, once the trained model is deployed, the data generally contains very few faulty images. This difference between the training set and the production set, which is similar to the test set, is addressed in~\cite{zhaoActiveLearningLabel2021}. The authors propose a framework that characterizes label shift across three cases, defined by the distributions of the selected (labeled) data, the unlabelled pool, and the inference (test) data. In this paper, we analyze multiple combination of distribution for pool and test distribution in order to measure the classical AL setting where pool and test distribution are identical, and situation that are closer to industrial situations where upstream rules determine which images are stored (to optimize storage), thereby shaping the unlabelled pool.
    In industrial settings, the underlying data distribution is unknown before labeling. Therefore, we empirically assess whether, under varying levels of class imbalance and label shift, standard AL strategies perform no worse than random sampling, we stress that this is an empirical observation supported by significance testing (Section~\ref{sec-results}), not a theoretical guarantee.

    \section{Methodology}\label{sec-method}
    
    \subsection{Active learning and semantic segmentation in machine vision}
    
    Pool-based AL applied to semantic segmentation in machine vision consists in  sequentially labeling a small amount of images extracted from an unlabelled dataset and then train a model based on this subsampled dataset. Let $D^u=\{X_i^u, i \in \mathcal D^u\}$ be the set of unlabelled images, where $X_i^u \in \mathbb R^{3q}$ are independent RGB images, where $q$ is the number of pixels. It is worth mentioning that $D^u$ could already be a subset of $D^c = \{X_i,i \in \mathcal D^c\}$ a set of collected (or acquired) images. Indeed, it is quite common in industrial setting that $\# \mathcal D^c$ is so massive that all images cannot be stored and that a preliminary filtering, most often depending on $X_i$, is performed. In other words, $\mathcal D^u$ is to be seen as a random discrete subset of $\mathcal D^c$ and $\mathrm P(i \in \mathcal D^u)$ depends on $i$. Now, that $D^u$ is defined, we assume that each image $X_i$ can be submitted to an oracle or expert who labels the image $X_i^u$, i.e. defines a mask $Y_i^u \in \{0,1\}^{3q}$ for any $i\in \mathcal D^u$. We further define the random variables $I_i^u =\mathbf 1(\|Y_i^u\|_1>0)$ where $\|z\|_1$ stands for the $\ell^1$-norm of $z\in \mathbb R^{3q}$. The variable $I_i^u$ indicates whether a given segmentation mask contains at least one pixel with class 1, i.e. the fact that an image is classified as one containing at least one defect or not. Given $\mathcal D^u$, the unlabelled dataset $D^u$ is characterized by $\pi^u =  \sum_{i\in \mathcal D^u} \mathrm P( I_i^u=1 \mid  i \in \mathcal D^u) / \# \mathcal D^u$ which represents the average proportion of faulty images among $D^u$.
    
    The AL proceeds as follows. At cycle $j=0,\dots,J$, we extract a set of images $S_j^u$ from the unlabelled dataset, i.e. a set of indices $\mathcal S_j^u \subset \mathcal D^u$ with size $\# \mathcal S_j^u=B$ called the budget. We form the set $S_j^u=\{X_i^u, i \in \mathcal S_j^u\}$. These images are labeled by the expert and the whole is stored in $S_j^\ell=\{(X_i^u,Y_i^u), i\in \mathcal S_j^u\}$. The model is then trained on $D_j^{\ell}=D_{j-1}^\ell \cup S_j^\ell$ (with $D_{-1}^\ell=\emptyset$). The trained model at cycle $j$, say $\hat f_j$, is directly used to score/predict every image from $D^u \setminus \cup_{k=0}^j S_k^u$. The scores or predicted probabilities are denoted by $s_{i,j}^u$ and {$\hat{ p}_{i,j}^u$}  for any $i \in \cup_{k=0}^j   \mathcal S_j^u$. The predicted probabilities or scores (depending on the AL strategy) are then used to extract the next set of images $S_{j+1}^u$ from $D^u \setminus \cup_{k=0}^j S_k^u$. The procedure, initialized with $\mathcal S_0^u$ randomly selected from $\{1,\dots,\#\mathcal D^u\}$, is performed $J$ cycles. 
    
\begin{figure}
    \centering{
        \includegraphics[width=\textwidth]{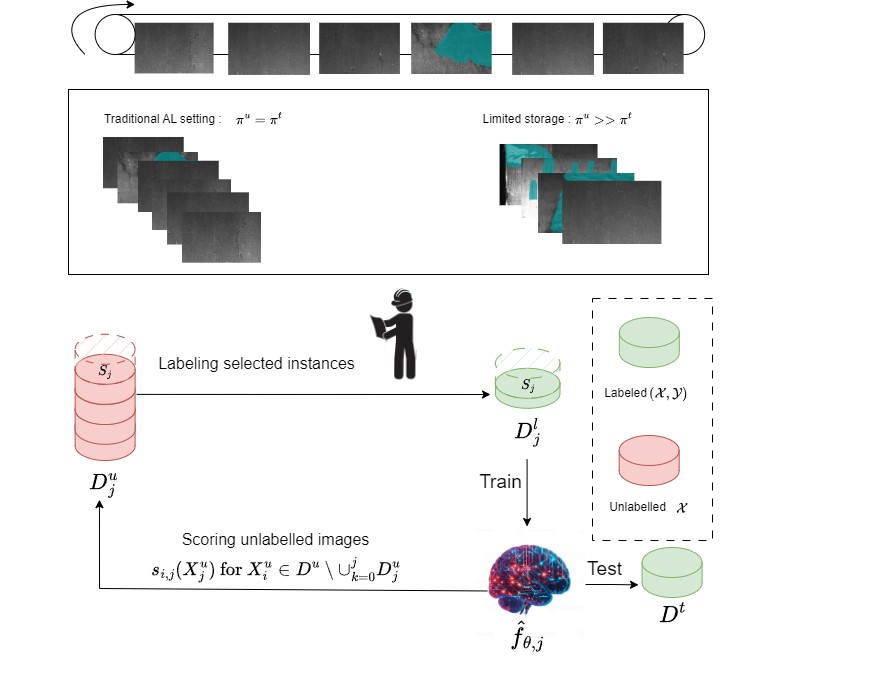}
    }
    \caption{From top to bottom. (i) Data acquisition: images are collected in the factory according to a production distribution $\pi^t$, (ii) The standard AL setting corresponds to the one where all collected images are stored (situation where $\pi^u=\pi^t$). Due to limited storage capacity, it may happen that only some images are stored according to expert rules (situation $\pi^u>>\pi^t$). (iii) Active Learning procedure: each cycle $j$ selects a pool of images, $S_j^u$, from $D^u$, label selected images (giving rise to $S_j^\ell$), train a model on on $D_j^\ell = D_{j-1}^\ell \cup S_j^\ell$ and uses this trained model $\hat{f}_{\theta,j}$ to select most informative images from $D^u\setminus \cup_{k=0}^j D_k^u$ with an acquisition (or score) function $s$.}
    \label{alschema}
    \end{figure}

    To track the performances of the AL algorithm, we apply the procedure on a test dataset $D^t=\{(X_i^t,Y_i^t), i \in \mathcal D^t\}$. In the industrial context, the test dataset is to be seen as representing model deployment, where the model is applied to real images acquired over time. Thus, in some sense, the way images are included in $D^t$ is the same than the initial ones from $D^c$, i.e. unlike the set $\mathcal D^u$, there is no reason that $\mathrm P(i \in \mathcal D^t)$ depends on $i$. The dataset $D^t$ is characterized by $\pi^t= \sum_{i\in \mathcal D^t}\mathrm P(I_i^t=1) / \# \mathcal D^t$ which measures the average proportion of faulty images among $D^t$. The whole procedure is illustrated in Figure~\ref{alschema}
    
    The next section details the simulation setting we consider. We mainly play with the two parameters $\pi^u$ and $\pi^t$. Since we do not model the distribution of $Y_i \mid X_i$, these probabilities are non-evaluable and we replace them by their empirical versions $({\mathcal D}^\bullet)^{-1} \sum_{i\in {\mathcal D}^\bullet} \mathbf 1(I_i^\bullet=1)$ for $\bullet=u,t$. We abuse notation by still denoting $\pi^u$ and $\pi^t$ these quantities. A small value of $\pi^u$ is related to the concept of {\bf class imbalance}, i.e. $D^u$ contains very few faulty images. The situation $\pi^t =\pi^u$ corresponds to a situation where the test and unlabelled datasets have the same distribution, while the situation $\pi^u>>\pi^t$ reflects that we have more faulty images in the unlabelled dataset than in the test one and illustrates a particular case of the concept of {\bf label shift} due to limited storage capacity, where the difference in distribution between the training set (i.e. unlabelled dataset) and the test set ensues from a limited storage constraint which imposes a pre-filtering of $D^c$. We emphasise that class imbalance and label shift are distinct notions: the former concerns the skew of the class marginal within a single set (a small $\pi^u$), whereas the latter is a change of this marginal between the pool and the test set ($\pi^u\neq\pi^t$) while the class-conditional distribution $\mathrm P(Y\mid X)$ is assumed unchanged the classical prior-probability shift.
    
    \subsection{Active Learning Algorithms}
    
    The focus of this paper is more on the sensitivity of active learning algorithms to class imbalance and label shift than the performances of the algorithm themselves. Hence, we aim to compare four state of the art acquisition functions used in semantic segmentation (\cite{mittalBestPracticesActive2024}) which are the entropy selection \cite{joshiMulticlassActiveLearning2009}, the core-set selection~\cite{senerActiveLearningConvolutional2018a} and galaxy sampling \cite{zhangGALAXYGraphbasedActive2022}. We also consider the basic strategy consisting in uniform random sampling.
    
    \paragraph{Random Sampling.} This strategy is the baseline, where we sample images from the unlabelled dataset according to a uniform distribution.
    
\begin{figure}
    \centering{
        \includegraphics[width=\textwidth]{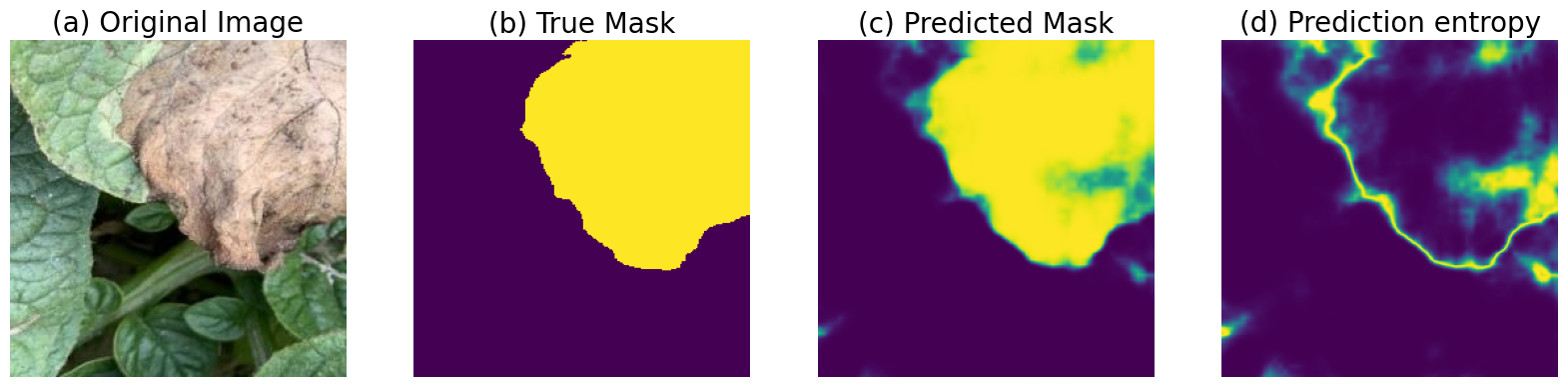}
    }
    \caption{(a) Example on an image from the patched potato disease dataset. (b) True faulty pixels in yellow. (c) Predicted probabilities for the defect class. (d) Pixelwise Entropy on the prediction of a trained model. Yellow : Faulty, Purple : Healthy. This is done with a model trained on the full patched dataset for illustration purposes.}
    \label{exentropy}
    \end{figure}

    \paragraph{Entropy Sampling.} Semantic segmentation being a collection of classifiers, we can compute the Shannon entropy for each predicted pixel and average them over the full image to get a level of uncertainty for this image. Let $\hat p$ be the vectorized predicted probabilities. The average image entropy, in the case of binary semantic segmentation, is then given by $q^{-1} \{ 
    \hat p^\top  \log \hat p + 
    (1- \hat p)^\top  \log (1-\hat p)
    \}$.
This strategy is representative of latent-space exploitation, in which the selection process specifically targets the weaknesses of the model. This method, referred to as entropy in the plots, aims to optimize the learning process by strategically choosing samples that are the most difficult for the model. This approach can improve model performance by concentrating on regions of high uncertainty. However, it also has limitations: it may overfit to specific weaknesses and overlook potentially valuable information in areas where the model already performs well. In the context of semantic segmentation in particular, averaging entropy values tends to favour images containing larger defects, as illustrated in Figure \ref{exentropy}.
    
\paragraph{Core-Set Sampling.} For the core-set selection, the algorithm proposed by \cite{senerActiveLearningConvolutional2018a} is considered by using the original Euclidean distance. The embeddings used for the core-set selection are the 
$L^2$-normalized features obtained from the global average pooling of the six multi-scale feature maps produced by the FPN encoder. This strategy focuses on {\it exploration} of the latent space, aiming to generate images that are highly diverse from one another. The method—referred to as deepcoreset in the next section—builds on the original DeepCore algorithm. By prioritizing diversity in the selected samples, we seek to enhance the model’s capacity to capture a broader range of features and characteristics within the latent space. While this approach can lead to a more comprehensive representation of the dataset, it may also introduce redundancy when the selected diverse samples overlap with regions that are already well understood.

\com{
\paragraph{Galaxy Sampling.} This strategy is adapted from \cite{zhangGALAXYGraphbasedActive2022}, with some modifications to make it suitable for semantic segmentation problems. The main characteristic of distribution-aware active learning acquisition functions in classification is that they estimate the labels from the model predictions and perform the selection based on the predicted distribution.
The strategy starts by ordering the prediction scores over the entire dataset ($D^\ell \cup D^u$). It then selects unlabeled items whose prediction scores lie between two consecutive labeled items belonging to opposite classes. If multiple pairs of labeled items belong to opposite classes, the idea is to identify unlabeled items that are as close as possible to the decision boundary. We therefore select the pair with the smallest gap.
}

\com{
In semantic segmentation, each image can (and usually does) contain multiple labels. To apply the Galaxy strategy to semantic segmentation, we estimate from the model output the probability that an image is faulty, defined as the average predicted probability over all pixels for the faulty class. We then apply the Galaxy strategy in a binary classification setting.
}

\subsection{Implementation details}

For this numerical study we consider the simple model for semantic segmentation which is a Feature Pyramid Network \cite{linFeaturePyramidNetworks2017a}. This neural network model is able to extract features of different sizes in images and is pre-trained with the weights of imageNet.
The problem is binary semantic segmentation, since each pixel can either be background or defect. The output of the model is a matrix with two channels where each channel represents the probability of each class of being true according to the model.
All models are optimized on NVIDIA T4 GPUS with AdamW (\cite{loshchilovDecoupledWeightDecay2018}) a batch size of 32 during 5 epochs with a learning rate of $5 \times 10^{-4}$. The loss function we consider is the Combo Loss \cite{taghanakiComboLossHandling2019}, combination of DiceLoss \cite{sudreGeneralisedDiceOverlap2017} and Binary Cross entropy Loss computed at the pixel-level and averaged over the whole image. Using multiple losses is quite common in semantic segmentation, see \cite{jadonSurveyLossFunctions2020}. Let $Y$ and $\hat p$ be respectively the vectorized binary mask and predicted probabilities of an image. This loss is given by
\begin{equation*}
\mathcal{L}_{Combo} = \frac{-2 \hat p^\top Y +S}{(\hat p + Y)^\top 1 +S} 
- \frac1q \;  \hat p^\top \log(\hat p) +(1-\hat p)^\top \log(1-\hat p)
\end{equation*}
where $S \in \mathbb{R}^+$ is a real number used for computational stability, set to 1, as recommended by~\cite{taghanakiComboLossHandling2019}.

\subsection{Evaluation Criteria}

To understand how AL strategies are affected by class imbalance and/or label shift, we intend to measure the efficiency and the variability of each AL strategy and to understand which type of images from the unlabelled datasets are selected at each cycle.
For the first task we propose  to consider \textbf{F1-Score of the faulty class} computed on the test sets. This metric evaluates the model’s prediction quality by considering both precision and recall, thereby providing a balance between the two in scenarios where the class distribution is skewed. We report the defect-class F1 as our primary metric, for binary masks it coincides exactly with the defect-class Dice coefficient, and the defect-class IoU is its strictly monotone transform $\mathrm{IoU}=\mathrm{F1}/(2-\mathrm{F1})$, so these three metrics rank strategies identically. For completeness we additionally report the global Mean IoU, in Section~\ref{sec-results}, the class-averaged Mean IoU is dominated by the (easy, majority) background class and is therefore less discriminative for defect segmentation. To measure the variability, we repeat the whole procedure {\bf 15 times}, restarting each time with a set $S_0^\ell$ randomly chosen from $D^u$. Additionally we evaluate the {\bf proportion of faulty images} selected by each AL strategy and for each cycle. In addition, to have an idea of which images are selected over the 15 repetitions, we construct a score called  {\bf uniqueness score} for each cycle, denoted by $\mathrm{us}_j$ and given by
\begin{equation}
\mathrm{us}_j = \frac{| \cap_{r=0}^R S_{j,r}^\ell| -b}{b(R- 1)}
\label{eq:us}
\end{equation}
 where $b$ is the number of images added at each cycle, $R=15$ is the number of repetitions of the AL procedure per cycle and where $S_{j,r}^\ell$ is the set of images selected and labelled at cycle $j$ for the repetition $r$. For a given cycle, the score equals 1 (resp. 0) when all selected images are different (resp. similar). \com{The uniqueness score is a purely descriptive diagnostic of selection overlap contrasting the exploration and exploitation behaviour of the strategies and is not intended as a measure of, or a bound on, model generalisation.}

\com{
\paragraph{Statistical significance.} As each configuration is repeated 15 times with paired initial sets, the differences between the strategies are assessed using a Welch test at the final cycle. A series of tests are being conducted for each acquisition function, as well as for each pair of values, i.e. ($\pi^u,  \pi^t$). We consider a false dicovery rate control and adjust the p-values (separately for each dataset) using the standard Benjamini-Hochberg procedure.
}
\section{Experiments and Results}\label{sec-results}

\subsection{Imbalanced dataset construction} 

In real-world scenarios, where the parameters of business rules and production distribution are not clearly defined, effective control over the distribution of unlabelled and test data is paramount. This overarching objective constitutes the foundation of the methodology employed in this study, which has not yet been utilised. To increase robustness of findings, we propose to consider two different open-source datasets. The first one is the {\bf potato diseases dataset} \cite{anupPotato_diseaseDataset2024} which contains agricultural images of potato leaves. It consists of 5165 3x512x512 RGB images of potato leaves. Each image is accompanied with a segmentation mask that we binarize, where the value 0 means healthy. Given that images are the primary data units rather than individual pixels, we categorize them into 'healthy images' when they contain no faulty pixels, and 'faulty images' in the opposite case. Note that all masks contain at least one non-zero pixel. The second considered dataset is the \textbf{Severstal dataset} \cite{severstal-steel-defect-detection} which is an open machine vision dataset with semantic segmentation labels of defects on metal plates. This dataset contains 18,074 256x1600 grey-scale images with 6,640 faulty images (37\%), the targets are binarized as well.

\begin{figure}
\centering

\begin{subfigure}[b]{0.2\textwidth}
  \includegraphics[width=\textwidth]{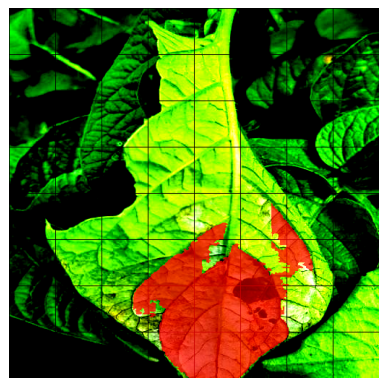}
  \caption{}
\end{subfigure}
\hfill
\begin{subfigure}[b]{0.15\textwidth}
  \includegraphics[width=\textwidth]{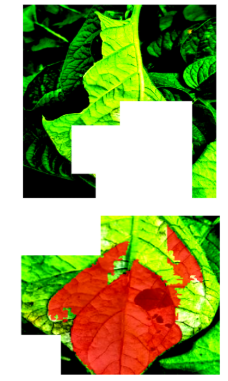}
  \caption{}
\end{subfigure}
\hfill
\begin{subfigure}[b]{0.25\textwidth}
  \includegraphics[width=\textwidth]{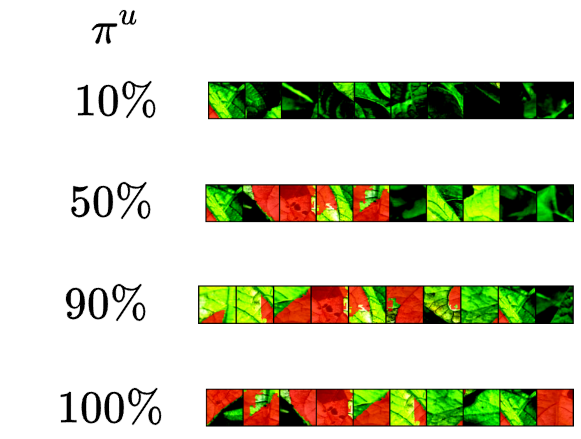}
  \caption{}
\end{subfigure}
\hfill
\begin{subfigure}[b]{0.2\textwidth}
  \includegraphics[width=\textwidth]{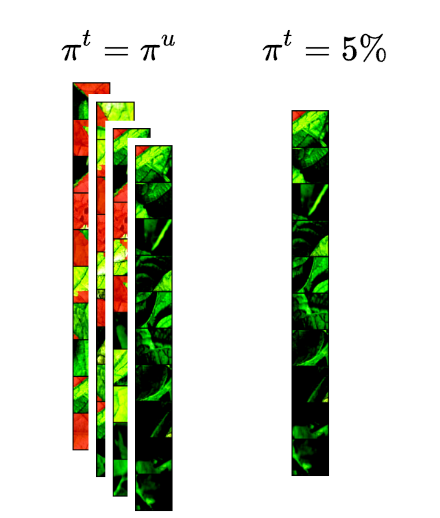}
  \caption{}
\end{subfigure}

\caption{(a) Original Image 512x512, Every image contains defective pixels (b) Patchification into 256x256 patches. This creates two sets of images : the first one composed of the healthy pixels and the second one with the defects only (c) Synthetic Class imbalance by randomly sampling each set of images according to a predefined proportion of defects (d) Test set distributions under AL assumption and industrial case fixed at 5\%}
\label{fig-patchification}
\end{figure}%

\com{In order to create datasets with different values of $\pi^t$ and $\pi^u$ and make the findings comparable between the datasets, we decide not to work on the full images but on patches of 3x256x256. After patchifying the potato diseases training dataset we end up with 25,788 smaller images. 20,341 of them are faulty and 5,447 are healthy, which is about 79\% of faulty images. The version of this dataset is named {\bf patched potato diseases}.}

\com{The patching algorithm may modify spatial correlation and exert an effect on the training process. To mitigate this potential effect, we have also created healthy and faulty samples in an additional manner, namely through resizing. The images depicting potato diseases were resized from 512x512 to 256x256, and images with a mask having a small volume were considered to be healthy. Small volume being defined as images having a volume smaller or equal to the quantile 0.21 of all volumes, which corresponds to the proportion of healthy images in the patched potato diseases dataset. Whilst it may introduce label noise during the training process, a parallel can be drawn with the process of patching. This version of the potato diseases dataset is referred to as the {\bf resized potato diseases}.} 

\com{The Severstal dataset contains greyscale image with dimension 256x1600, we patch each image into 256x400 smaller images. We end up with 43,048 patches with 9,358 faulty ones. (22\%). This version of the Severstal dataset is referred to  {\bf patched Severstal diseases}.}

\com{Let us provide some more details on the construction of the two patched datasets. Since we aim to construct subsets spanning $\pi^u$ from $0\%$ to $100\%$ while keeping the pool size fixed, the binding resource is the minority category: the $5,447$ healthy potato patches. (The $20,341$ faulty and $25,788$ total potato patches, and the $43,048$ Severstal patches, refer to the full inventories). We therefore cap the unlabelled pool at $4,300\approx 79\%$ of $5,447$, leaving $\approx 1,147$ healthy patches to build the mostly-healthy ($\pi^t=5\%$) test set.} The datasets are constructed randomly by sampling the required number of items from each training or test set to reach any desired value of  $\pi^u$ and $\pi^t$ (see Figure~\ref{fig-patchification} with one image as an example). \com{To investigate the class imbalance sensitivity and label shift sensitivity, we let $\pi^u, \pi^t \in \{5\%, 25\%,50\%,75\%,95\%,100\% \}$. The label shift sensitivity is actually controlled by the difference between $\pi^u$ and $\pi^t$.} 
We run this experiment for $J=10$ cycles, so that $\# \mathcal D_j^u /\# \mathcal D^u=2\%,4\%,\dots,20\%$. Each cycle is repeated 15 times. We report the F1-score with uncertainty bands represented by the interquartile range. Figures~\ref{fig-perf_f1faulty} summarises the main results. 

\subsection{Results}\label{sec-resultss}

\begin{figure}
\centering
\includegraphics[width=\textwidth]{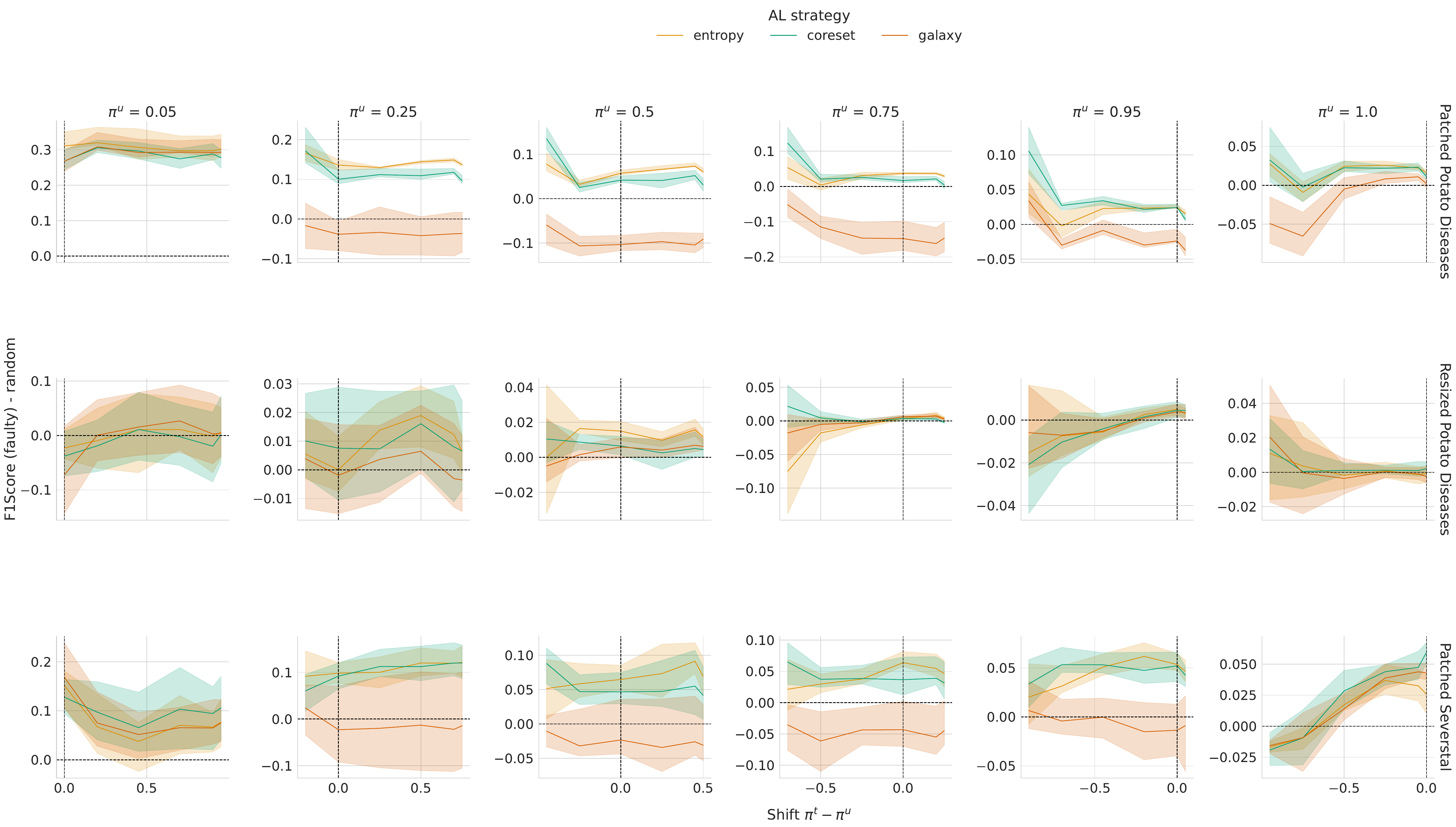}
\caption{Difference of f1 score for the faulty class of each AL strategies (colors) against random sampling, separated for each datasets (rows) and each values of $\pi^u$ (rows). Curves above zero means that AL beats random sampling. x-axis represents the shift between unlabelled pool distribution and test distribution, the lefter the more defect there are in the test set.
}
\label{fig-perf_f1faulty}
\end{figure}

\com{The main results of this work consist of complete AL performance along selection budget for each couple $(\pi^u, \pi^t)$ which is quite dense (see Figures \ref{fig-complete_potato_f1faulty},  \ref{fig:complete_resizedPotatof1faulty}, \ref{fig-complete_severstal_f1faulty} respectively for the patched potato diseases, resized potato diseases and patched severstal datasets). These figures represent the complete evolution of performance at each budget for each class imbalance value (columns) and each label shift (rows) where the diagonal depicts the no label shift situation ($\pi^u = \pi^t$). First of all, as expected, the higher the budget the higher F1-scores and this holds for all AL strategies. Entropy, coreset and galaxy based methods clearly outperform the naive random sampling for low values of $\pi^u$ and for most dataset, with different effect sizes. Figure \ref{fig-perf_f1faulty} summarizes this whole grid of plots for the three datasets.  For each dataset (rows) and each unlabelled pool composition $\pi^u$ (columns) it reports the difference between the defect-class F1 obtained by an AL strategy and the one obtained by random sampling, as a function of the shift $\pi^t-\pi^u$, at the final labelling budget ($860$ images, i.e.\ $20\%$ of the pool). A value above the dashed zero line means that, for the same annotation cost, the strategy trains a better model than random selection. We compare three active learning strategies and the multiple settings of distribution shifts with varying values of $\pi^u$ and $\pi^t$. It is reduced version computed by the performance gap with random selection for the last cycle.}

\com{We first comment Figure \ref{fig-perf_f1faulty} upper row (patched potato diseases dataset). It appears that the dominant is class imbalance and not label shift, indeed, the average gain of entropy sampling (resp core-set) on the potato diseases dataset decreases with $\pi^u$ going from $+0.344$ (resp $+0.324$) at $\pi^u= 5\%$ to $+0.008$ (resp $+0.015$) at $100\%$. The ordering is the same for the patched severstal dataset with smaller gains. The important effect size is observed only with $\pi^u=5\%$ where entropy and core-set sampling leads to a f1 score three times better than random sampling at the final cycle.
The second observation is the flatness of gains along the label shift axis. Indeed, on the patched potato disease dataset, the entropy and core-set curves remains stable for all shifts, patched severstal curves depicts the same patterns. It shows that benefits of AL does not cancel with pool/deployment distributions mismatch, however, it lowers the performance of every strategy, including random.
The galaxy strategy behaves qualitatively differently from the two other strategies. Its improvement is restricted to high imbalance situations and becomes detrimental as soon as the distribution start to become slighly balanced. This phenomemon could be explained by the fact that in pools richer in defects, items lying between predicted-faulty and predicted-healthy pairs are more likely to be healthy.}

\com{The results for the patched Severstal dataset, exhibit greater variability (as indicated by wider interquartile pointwise bands), and the overall performance is slightly lower than for the first dataset. This can be attributed to the higher complexity of the images under analysis. Nevertheless, most of the observations made for the patched potato diseases dataset still apply to the patched Severstal dataset, which, in our view, reinforces the usefulness and reproducibility of the present study.}

\com{
Finally, we want to see if the results differ depending on the way we corrupt the data. Indeed, patching breaks spatial correlation of pixels in each images. Patched potato diseases dataset being 512x512 square images, we corrupted the images by resizing them to 256x256 and considering images with mask's volume lower than 1284 pixels (21\% quantile, proportion of healthy to faulty created thought patching.) as healthy. This approach is questionable because all images actually contains faulty pixels, but the pixels coming from small masks are considered healthy so it adds some ambiguity to the supervision.
}

\begin{figure}
\begin{tabular}{|c|}
\hline
\includegraphics[width=\textwidth]{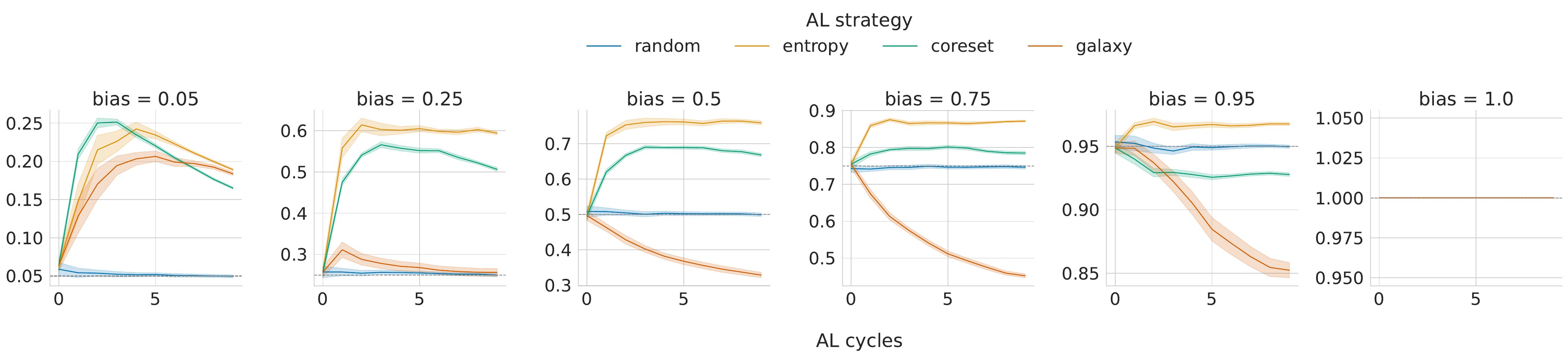}\\
(a) Cumulative proportion of faulty images selected over each AL cycles\\
\hline
\includegraphics[width=\textwidth]{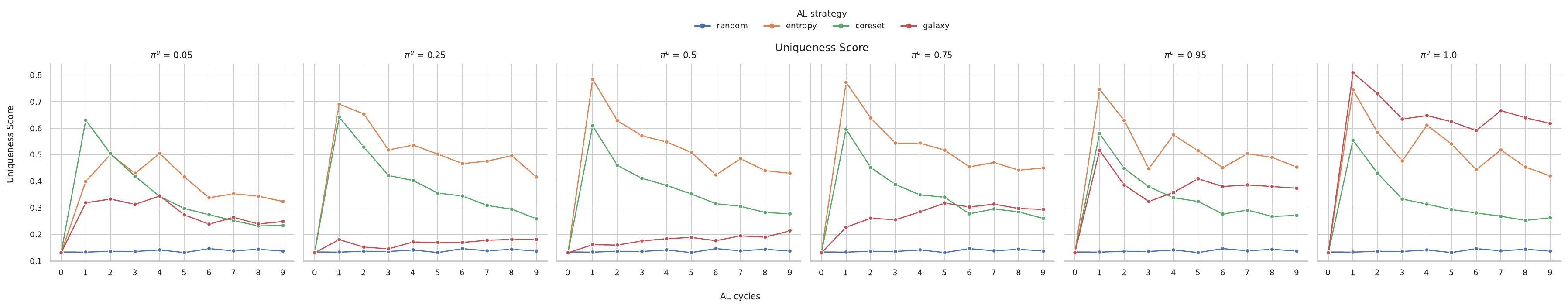}\\
(b) Uniqueness scores \\
\hline
\end{tabular}

\caption{
(a) Proportion of faulty images selected at each cycle for the \textbf{patched potato diseases} dataset. 
(b) Uniqueness scores for the patched potato disease dataset defined by~\eqref{eq:us}. Remind that a uniqueness score of 0.5 means that among all images selected at cycle $j$, half of them are unique.}
\label{fig-defectcum_potato} 
\end{figure}

\com{One may wonder which types of images are selected throughout the cycles of the different active learning (AL) strategies. Figure~\ref{fig-defectcum_potato} provides insights for the patched potato disease dataset, and similar observations hold for the two other datasets. Overall, Figure~\ref{fig-defectcum_potato} shows that, unlike random sampling, both the entropy-based and core-set strategies tend to select a higher proportion of faulty images than the underlying proportion $\pi^u$. This behavior is particularly pronounced for the entropy-based method, which selects on average 20\% faulty images, even though the unlabeled dataset contains only 5\% of such images. Although these two methods share this tendency, they differ in the specific images they select. The uniqueness scores obtained with the deep core-set strategy are much more variable than those of the entropy-based strategy, further illustrating the notions of exploration and exploitation associated with these two approaches, respectively.}


\com{
\subsection{Additional metrics}\label{sec-addmetrics}
Beyond the defect-class F1, Figure~\ref{fig-complete_potato_meaniou} reports the global Mean IoU and the per-class (defect) IoU on the patched potato diseases test set. As anticipated in Section~\ref{sec-method}, the class-averaged Mean IoU is dominated by the background class and varies little across strategies, so it discriminates poorly, the defect-class IoU, being the monotone transform $\mathrm{F1}/(2-\mathrm{F1})$ of the defect F1, reproduces exactly the ranking already visible in Figure~\ref{fig-perf_f1faulty}. 
\subsection{Qualitative examples}\label{sec-qualitative}
 As demonstrated in ~\ref{fig:qualitative_example}, the predictions for the defect class of a test image at each cycle are illustrated for models that have been trained according to each different acquisition function.
It is evident that models trained with random sampling encounter significant challenges in accurately predicting defect areas, if they do so at all. Conversely, AL strategies demonstrate a capacity to predict defect areas with increasing accuracy as the number of cycles increases, a finding that aligns with the metrics outcomes. 
}

\begin{figure}
    \centering
    \includegraphics[width=\textwidth]{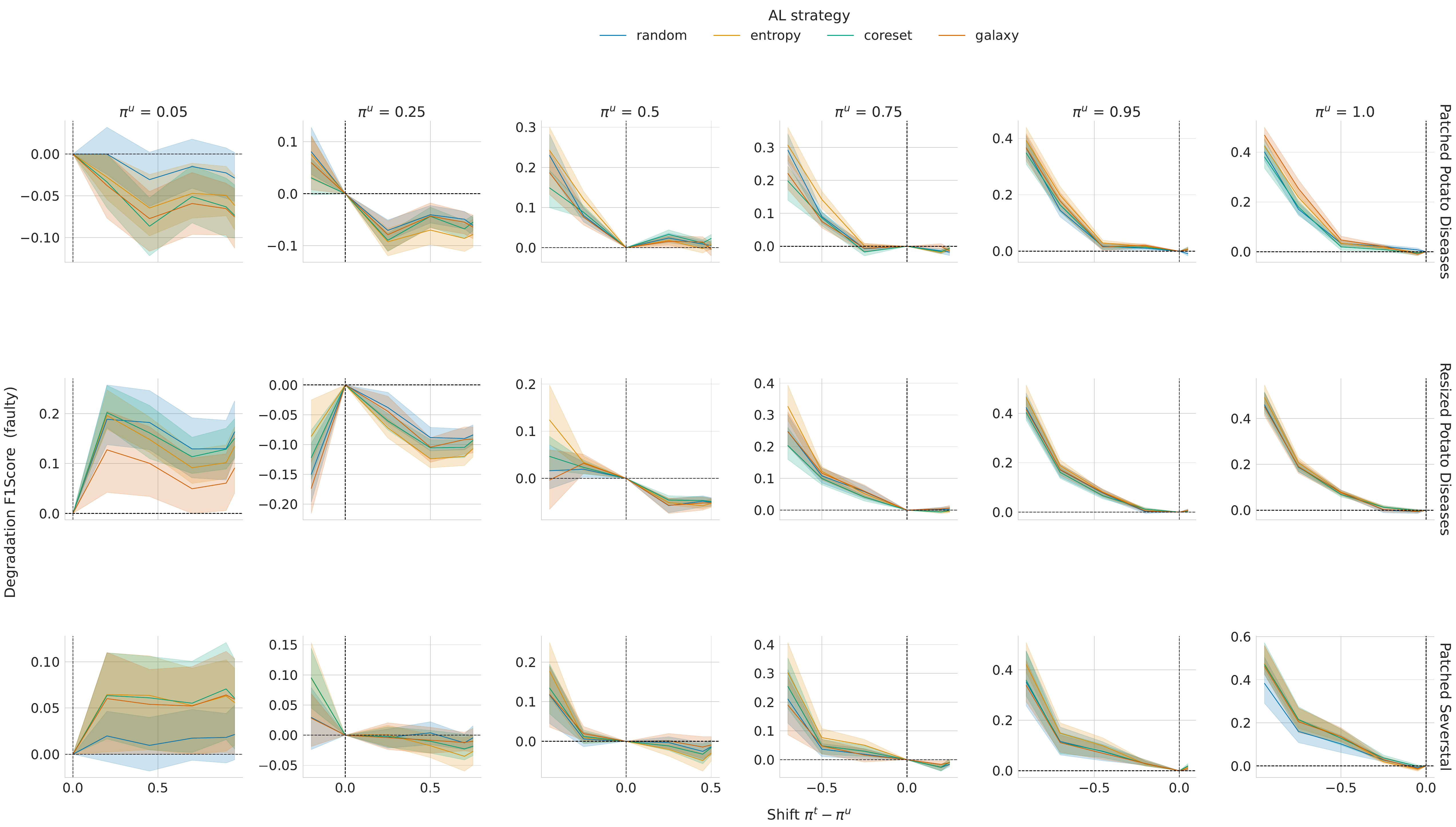}
    \caption{Degradation of the defect-class F1 caused by label shift, at the final labelling budget. For each strategy (colors) and each pool composition $\pi^u$ (columns), we measure $\mathrm{F1}(\pi^t=\pi^u)-\mathrm{F1}(\pi^t)$ evaluated on the same trained model, so that higher means worse generalisation and the curves pass through zero at $\pi^t=\pi^u$ by construction. Rows separate the two datasets, bands are interquartile range over the $15$ repetitions.}
    \label{fig-shift_f1faulty}
\end{figure}

\com{
\subsection{Sensitivity to the label-shift level}\label{sec-shiftcurve}
Because every model is now scored on the whole grid of test prevalences, we can characterise the loss due to label shift as a function of the shift magnitude instead of at a single contrasted point. Figure~\ref{fig-shift_f1faulty} plots, for each $(\pi^u,\text{strategy})$, the degradation $\mathrm{F1}(\pi^t=\pi^u)-\mathrm{F1}(\pi^t)$ of the same trained model against $\pi^t-\pi^u$, a positive value means worse generalisation than in the matched setting.
}
\com{
First, the degradation is strongly asymmetric. A positive shift (the deployment stream contains more defects than the pool) costs essentially nothing and is even mildly beneficial in some configurations, whereas a negative shift, which is precisely the storage-constrained industrial case, is very costly. Second, the cost grows with the mismatch and with $\pi^u$. Moving the deployment prevalence down to $\pi^t=5\%$ costs the entropy strategy up to $6.7\%$, $20.2\%$, $34.1\%$, $46.7\%$ and $51.6\%$ of its matched defect F1 on the patched potato dataset for $\pi^u=25\%,50\%,75\%,95\%$ and $100\%$. Most of the drop is already realised for a shift of about $-0.25$: the curves are convex and then flatten, so a moderate mismatch is comparatively benign while a large one is way worse. Third, the degradation curves of the four strategies are almost overlapped. Random sampling loses up to $8.4\%$, $19.5\%$, $35.1\%$, $46.0\%$ and $50.2\%$ on patched potato and $25.4\%$, $35.1\%$, $54.6\%$, $72.0\%$ and $79.8\%$ on patched Severstal for the same configurations, i.e.\ as much as, and on patched Severstal slightly more than entropy. Label shift is therefore a property of the deployment mismatch rather than a weakness specific to active learning: it lowers the whole ranking without reordering it, which is exactly why the differences of Figure~\ref{fig-perf_f1faulty} remain roughly constant along the shift axis. In absolute terms, however, the situation remains difficult: with $\pi^u=100\%$ and $\pi^t=5\%$ the defect F1 falls to $0.40$ (patched potato diseases) and $0.14$ (patched Severstal) whatever the strategy, so no acquisition function can compensate for a badly chosen business rule.
}

\com{
\subsection{Ablation: architecture and loss}\label{sec-ablation}
To check that our conclusions are not tied to a specific model, we repeat the core experiments with a different backbone (deeplabv3+ \cite{chenEncoderDecoderAtrousSeparable2018}) and a different loss (Focal loss \cite{linFocalLossDense2018}). The absolute F1 levels change with the configuration, but the qualitative ranking entropy outperforming random under no shift, most clearly at high imbalance and moderate budget, and this advantage narrowing under label shift is preserved.
}

\begin{figure}
    \centering
    \includegraphics[width=\textwidth]{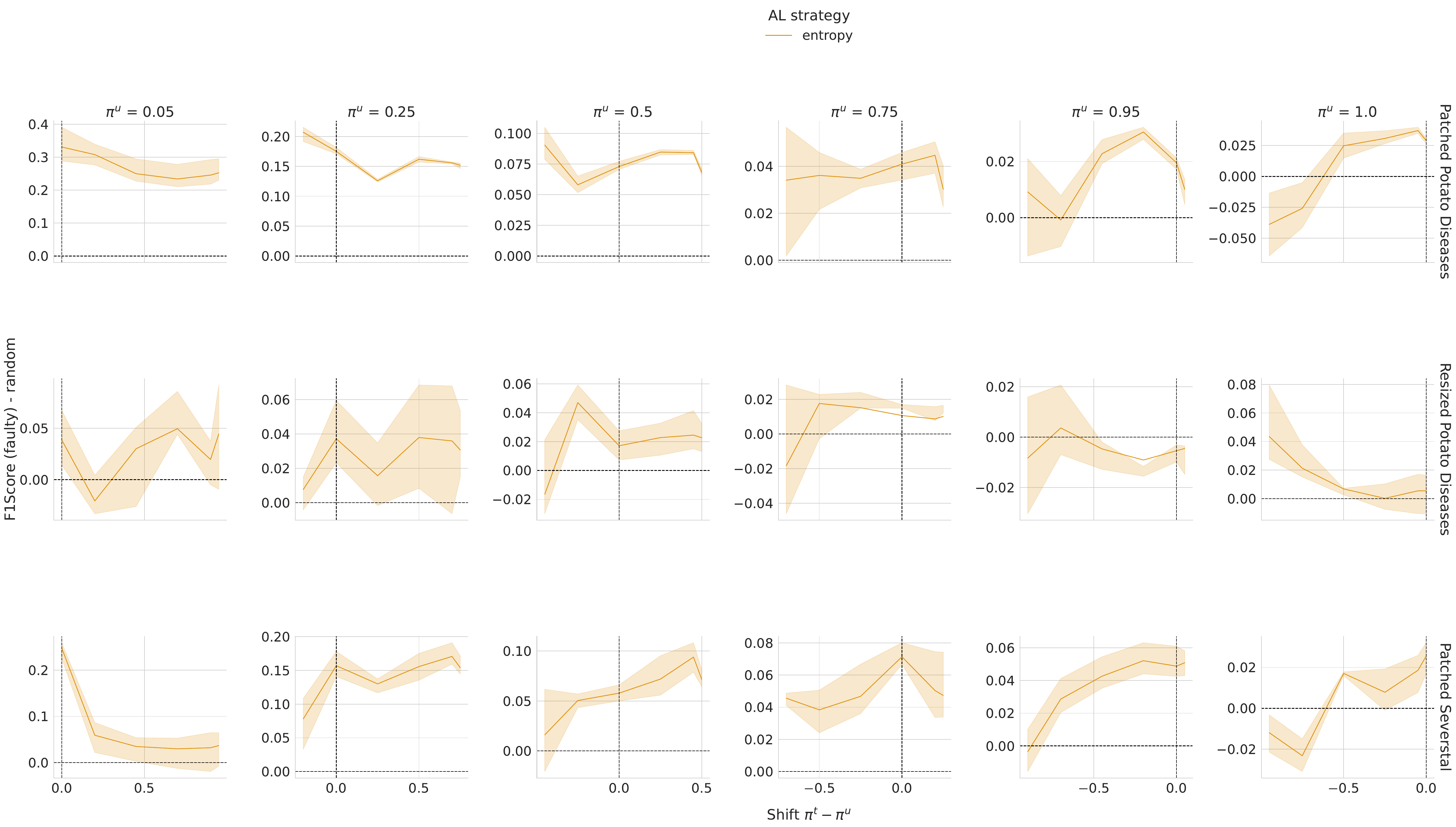}
    \caption{Experiment ran for ablation purposes on only 5 repetitions with one acquisition function (entropy sampling) - \textbf{Model : Deeplabv3+}}
    \label{fig:ablation_model}
\end{figure}

\begin{figure}
    \centering
    \includegraphics[width=\textwidth]{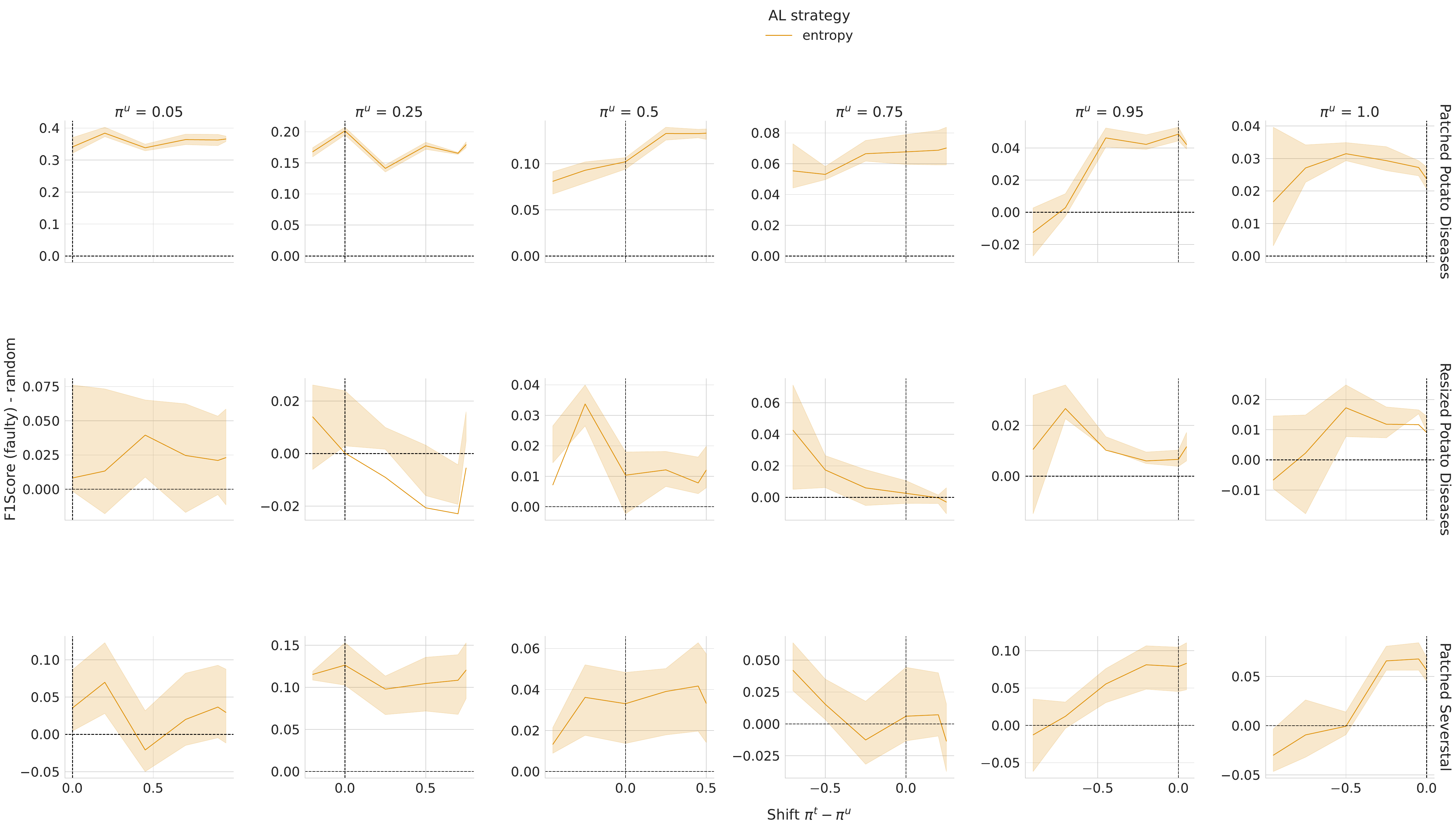}
    \caption{Experiment ran for ablation purposes on only 5 repetitions with one acquisition function (entropy sampling) -  \textbf{loss: Focal loss}}
    \label{fig:ablatoin_loss}
\end{figure}

\com{
\subsection{Statistical significance}\label{sec-signif}
The three datasets behave differently in the details, but two conclusions are common for the two patched ones. First, the sign of the significant differences separates the acquisition functions cleanly: wherever entropy or core-set differ significantly from random, the difference is in their favour, and there is no cell in which either of them is significantly worse than random, galaxy sampling, on the contrary, produces significant differences of the opposite sign in a substantial part of the grid, which confirms that its class-balancing rationale is actively harmful once the pool is no longer strongly imbalanced. Second, the significant cells concentrate at low $\pi^u$ and thin out as $\pi^u$ grows, in line with the reading of Section \ref{sec-resultss}: it is the imbalance of the unlabelled pool, not the pool/test mismatch, that determines whether active learning is measurably better than random. Moving along a row of the heatmap, i.e. varying $\pi^t$ at fixed $\pi^u$, does not reorganise the pattern, which is the statistical counterpart of the rank preservation observed in Section \ref{sec-shiftcurve}. Lastly, we can confirm what was visible on the \ref{fig-perf_f1faulty} where the resized potato diseases dataset shows no gain of AL over random sampling at the last cycle, however, it appears that on the early cycles, AL leads to better performances than random sampling when $\pi^u \in \{5\%, 10\%\}$ on Figure \ref{fig:complete_resizedPotatof1faulty}
}

\begin{figure}
    \centering
    \includegraphics[width=\textwidth]{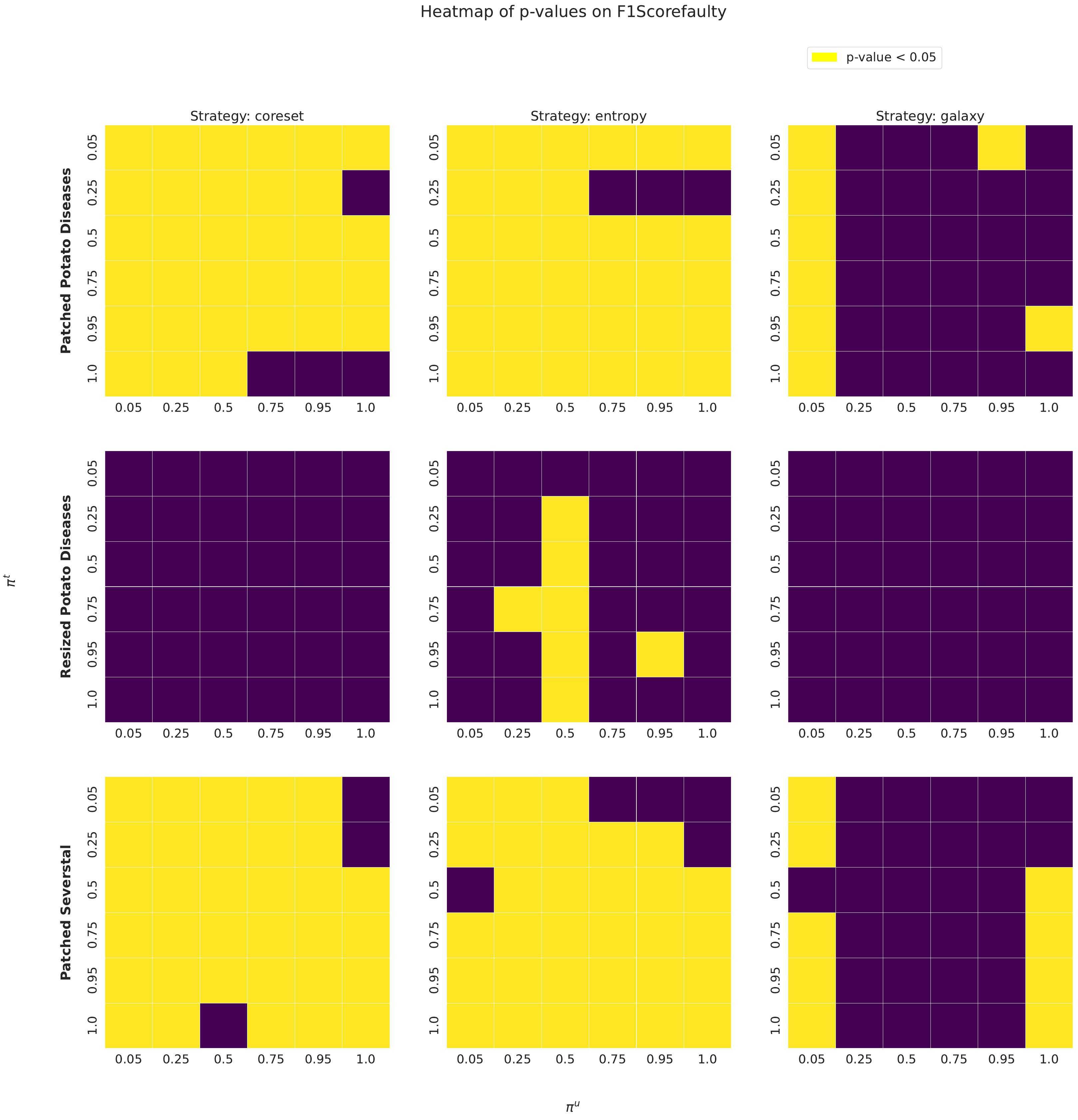}
    \caption{Heatmap representing wether the BH corrected p-values for the three datasets (rows) are less than the chosen threshold $\alpha = 5\%$. A yellow cell means that the corrected p-value for the test is less than the risk, and purple otherwise. }
    \label{fig:pvalueheatmap}
\end{figure}

\section{Discussion}\label{sec-discussion}

First, class imbalance and label shift affect two different quantities. Class imbalance ($\pi^u$) greatly affect the gain of AL over random image selection. The more imbalanced the data are, the greater the advantage of AL over random sampling. Label shift however, affect the absolute level reached by each strategy, but not the differences among them, keeping the rank unchanged. Second, degradation coming from label shift seems to be asymmetrical, indeed adding items containing defect does not deteriorate the performance of the model while the reverse, which is the situation created by the pre-filtering, is costly. Lastly, acquisition functions are not equal with unbalance regimes, the adapted galaxy sampling strategy is worse than random selection in half of the tested configurations. The potential causes of this phenomenon could be the class-balancing rationale, which results in an under-sampling of faulty images, or the adaptation methodology employed for segmentation segmentation, which may have been suboptimal.
 
Entropy selects the highest-uncertainty patches, which are disproportionately faulty because uncertainty concentrates on defect boundaries. At $\pi^u=5\%$ its cumulative selection reaches $24\%$ of faulty images, about five times the pool prevalence (Figure~\ref{fig-defectcum_potato}), so that after labelling $20\%$ of the pool it has recovered three quarters of all the defective patches. This is what makes it so effective under strong class imbalance. On more balanced pools the same mechanism biases the implicit class prior of the model towards the defect class. This bias does not translate into a larger sensitivity to label shift: the degradation curves of entropy and of random sampling are almost overlapped (Figure~\ref{fig-perf_f1faulty}).

\com{Our protocol controls the marginal proportion of faulty images, while patchification preserves the local spatial structure of each defect. However, independently drawing patches does not reproduce the inter-patch spatial correlation of defects across an entire part, nor does it preserve the empirical distribution of defect sizes and shapes. Moreover, the study is limited to two binary datasets, so we can only claim that the relative behaviour reported here holds for the configurations considered in our experiments.}

\com{Extension to multiclass segmentation, is not straightforward: with more than two classes a patch's label becomes a subset of classes, so a single scalar no longer parameterise the pool, and class co-occurrence within patches turns exact prevalence control into a constrained combinatorial resampling problem, feasible partial extensions include one-vs-rest control of a target class or stratification by the dominant class. On the contrary, when defect types are largely unknown a priori, unsupervised or semi-supervised anomaly detection trained on healthy images is an attractive complement for instance as a storage pre-filter or as an acquisition signal although it does not by itself deliver the pixel-level masks required here and carries its own calibration and false-positive challenges.}

\section{Conclusion}\label{sec-conclusion}

This controlled study shows that the amount by which active learning improves on random sampling is governed almost entirely by the class imbalance of the unlabelled pool, and very little by the pool/deployment mismatch. At $\pi^u=5\%$ entropy and core-set nearly triple the defect-class F1 at equal annotation cost and recover about three quarters of the defects present in the pool after labelling only $20\%$ of it. At $\pi^u\ge 95\%$ the gain becomes negligible, which is the regime where random sampling is a perfectly reasonable default. Label shift, by contrast, degrades all strategies by comparable amounts, asymmetrically (a defect-poor deployment stream is costly, a defect-rich one is not) and steeply, so that it must be addressed at the level of the storage/filtering policy or by prior correction rather than by the choice of an acquisition function. Beyond F1-scores, AL-trained models exhibit greater stability. Indeed, they are less dependent on the initial labeled set and rarely degrade as more labeled data are added. By systematically prioritizing informative samples from large unlabelled pools, AL mitigates the operational burden of annotating predominantly non-defective images and reduces sensitivity to unlabelled pool composition. Across the whole grid, entropy and core-set are never found significantly inferior to random sampling and are significantly superior in a large part of the configurations, whereas the distribution-aware galaxy criterion is significantly harmful as soon as the pool is not strongly imbalanced, as discussed in Section \ref{sec-signif}, the non-inferiority is an empirical finding at $R=15$ repetitions and not the outcome of an equivalence test. In conclusion, the AL approach provides a pragmatic and extensible framework for semantic segmentation in machine vision. Nevertheless, its efficacy appears to be constrained in scenarios characterised by significant class imbalance and a substantial amount of unlabelled data.

\section*{Declarations}

\subsection*{Data availability statement}
The two datasets used in this paper are open-source and can be found at the following URLs : 

The potato disease dataset can be downloaded here \url{https://universe.roboflow.com/anup-kaygm/potato_disease-binb3}.

The Severstal dataset can be downloaded here \url{https://www.kaggle.com/c/severstal-steel-defect-detection.}

The code is available on this repository : \url{https://github.com/JulienStats/WhenImbalanceComesTwice}

\subsection*{Competing Interests}
The authors declare that they have no Conflict of interest.

\bibliography{Patates}

\appendix

\newpage

\section{Patched potato diseases complete results plots}

\begin{figure}[h!]
    \centering
    \includegraphics[height=.8\textheight]{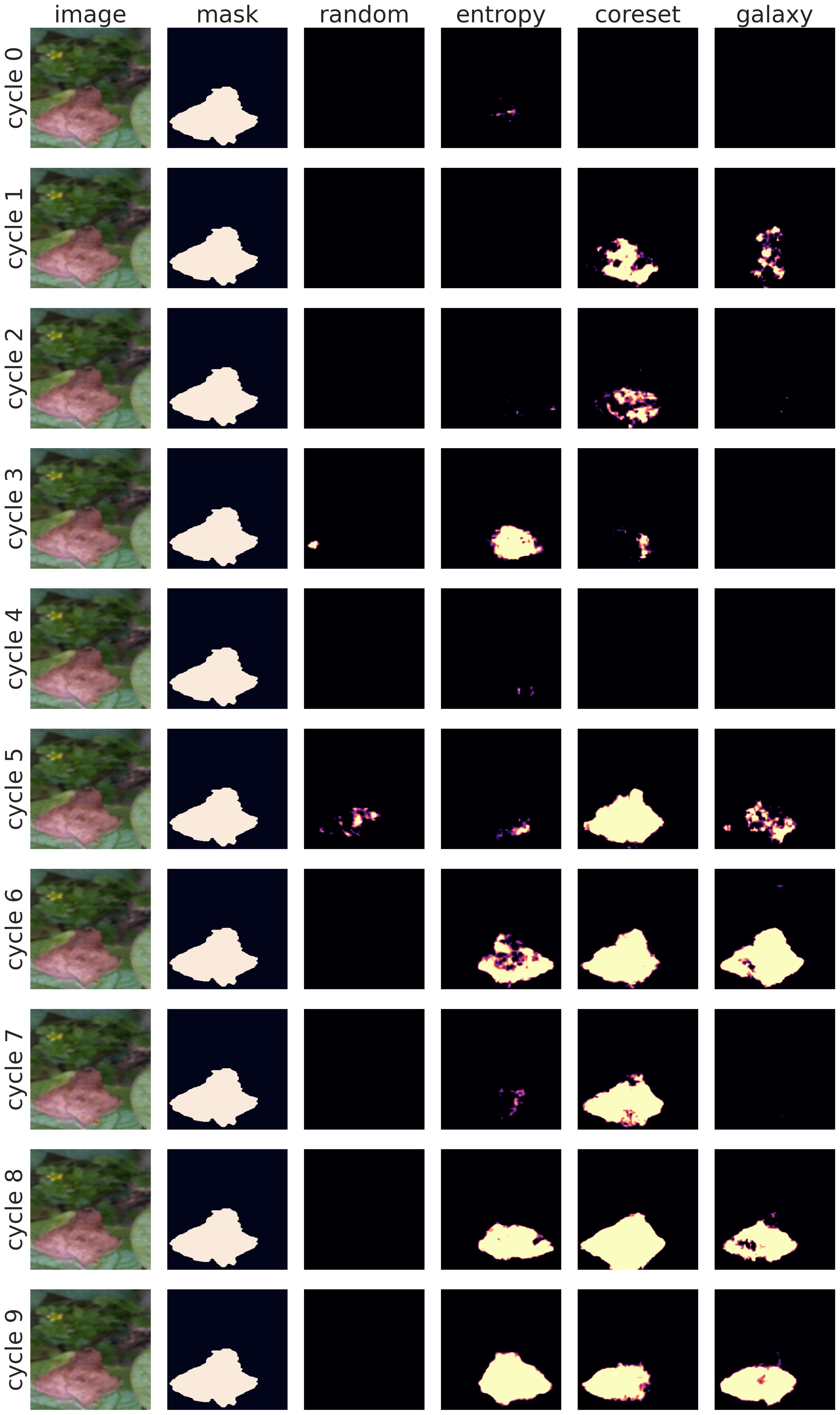}
    \caption{Qualitative example of prediction on a test image from the {\bf patched potato diseases} dataset for models trained at each cycle by each active learning strategy in the setting $\pi^u = 0.05$ }
    \label{fig:qualitative_example}
\end{figure}

\begin{figure}[h]
    \centering
    \includegraphics[width=\textwidth]{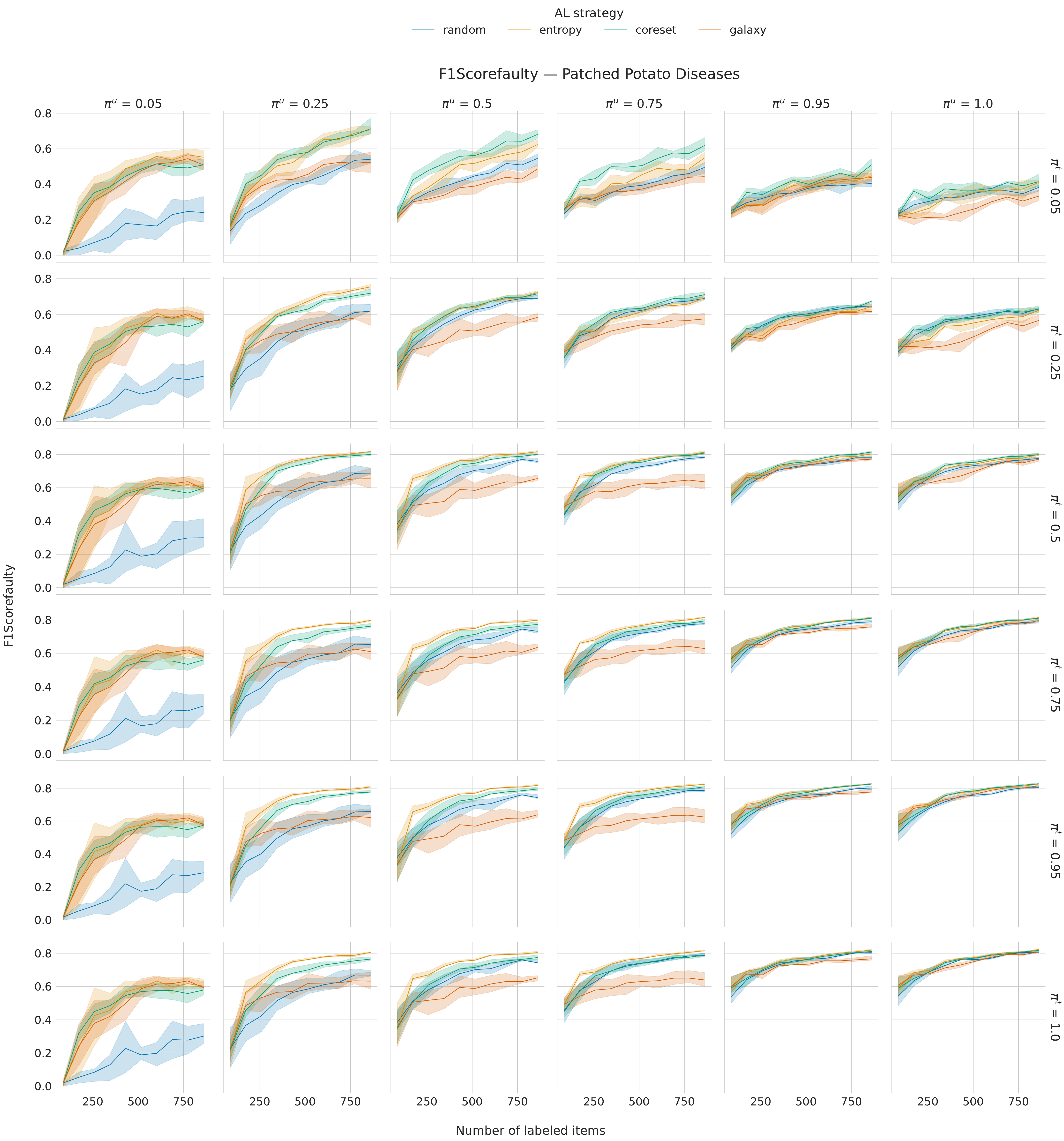}
    \caption{Complete AL results for each pairs of $\pi^u$ | $\pi^t$ measuring F1 Score of the faulty class for the \textbf{patched potato diseases} dataset. 
    Unit graphs depict the relation of F1 score for the faulty class along the labelling cycles. The global row axis represent the test set distributions. The column axis represent the distribution of the unlabelled dataset. The diagonal is the classical AL setting where $\pi^u = \pi^t$.}
    \label{fig-complete_potato_f1faulty}
\end{figure}

\begin{figure}[h]
    \centering
    \includegraphics[width=\textwidth]{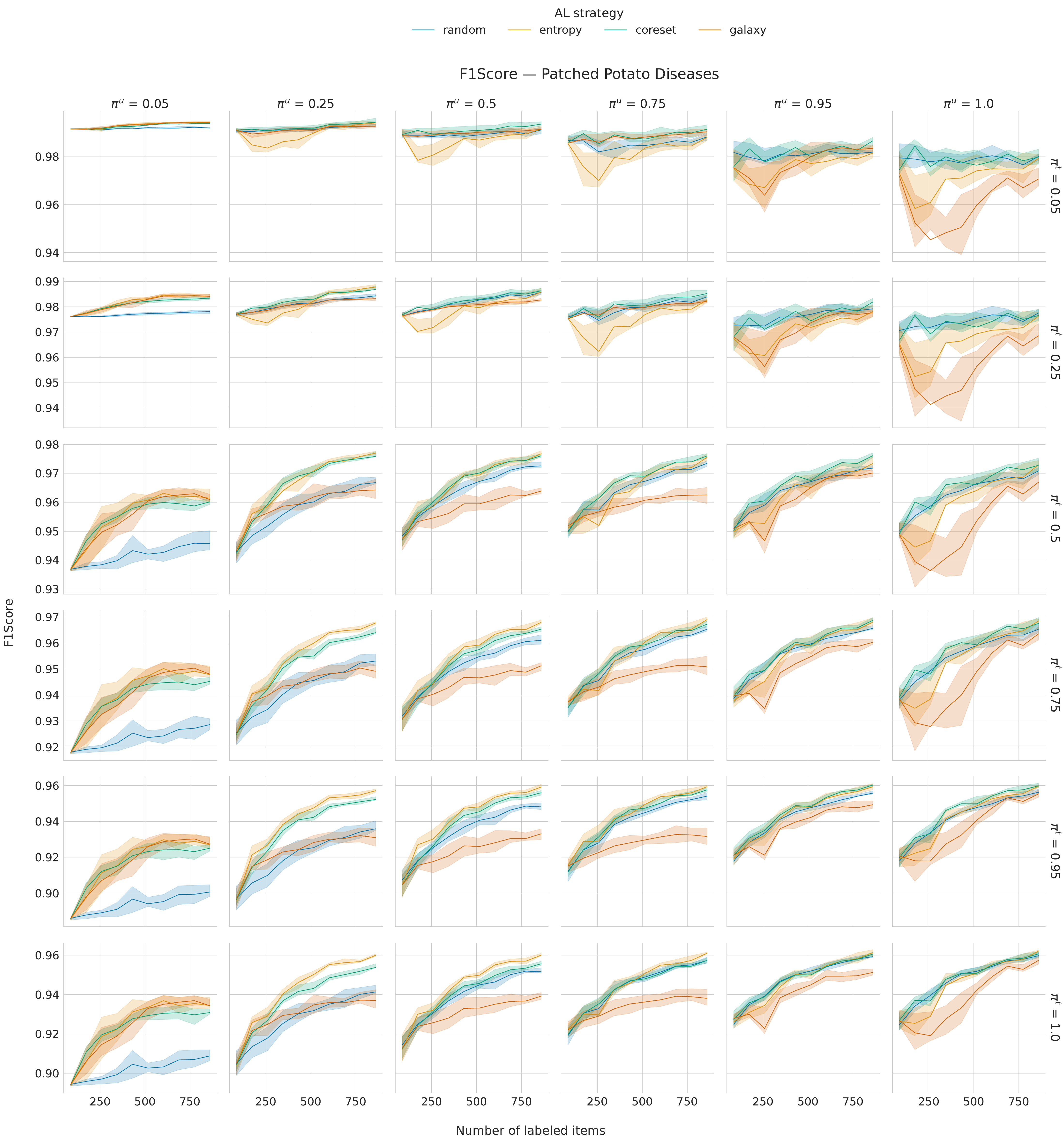}
    \caption{Complete AL results for each pairs of $\pi^u$ | $\pi^t$ measuring average F1 Score for the \textbf{patched potato diseases} dataset}
    \label{fig-complete_potato_f1}
\end{figure}

\begin{figure}[h]
    \centering
    \includegraphics[width=\textwidth]{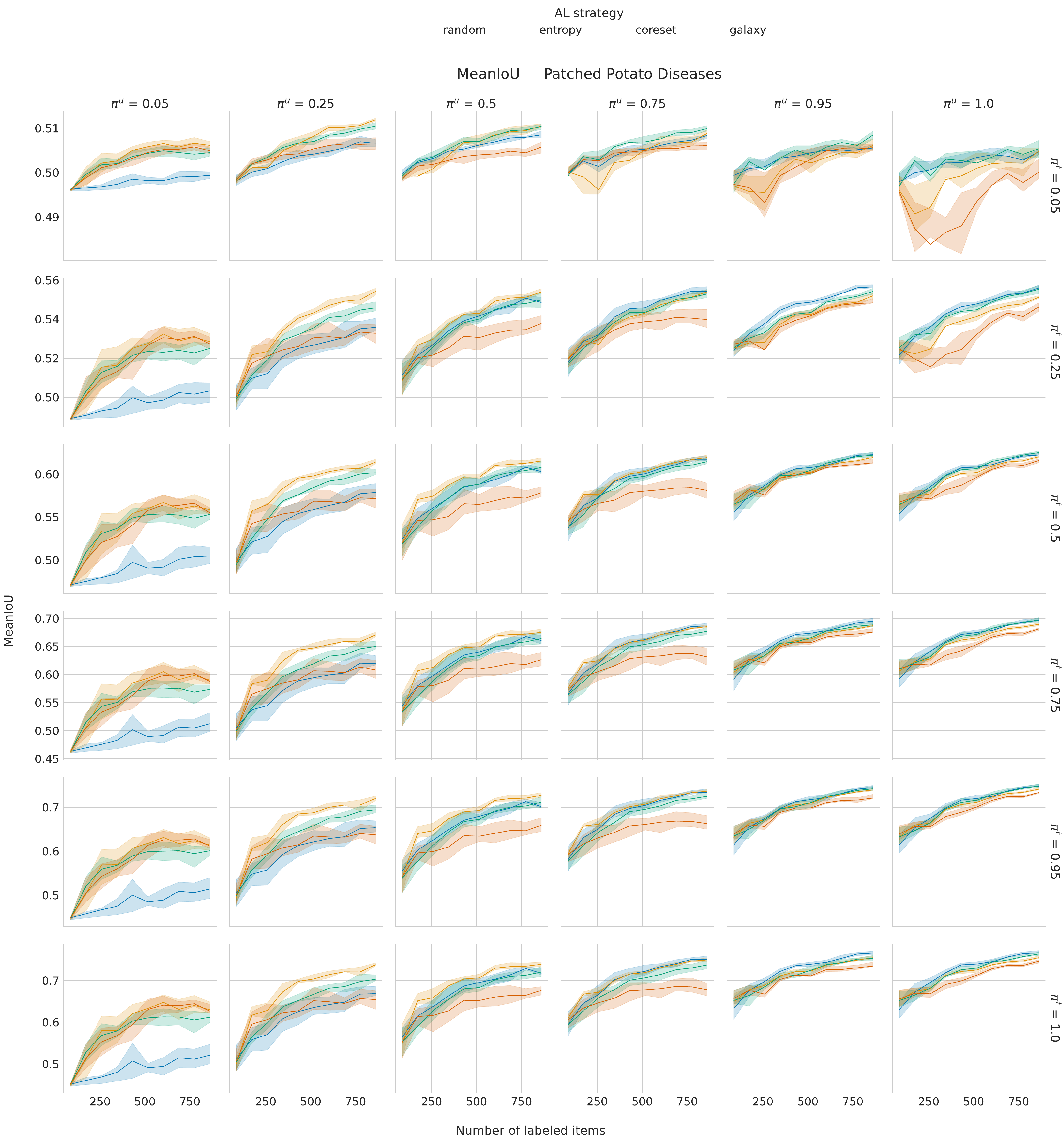}
    \caption{Complete AL results for each pairs of ($\pi^u$, $\pi^t$) measuring MeanIoU for the \textbf{patched potato diseases} dataset}
    \label{fig-complete_potato_meaniou}
\end{figure}

\newpage

\section{Resized potato diseases complete results plots}

\begin{figure}[h!]
    \centering
    \includegraphics[height=.8\textheight]{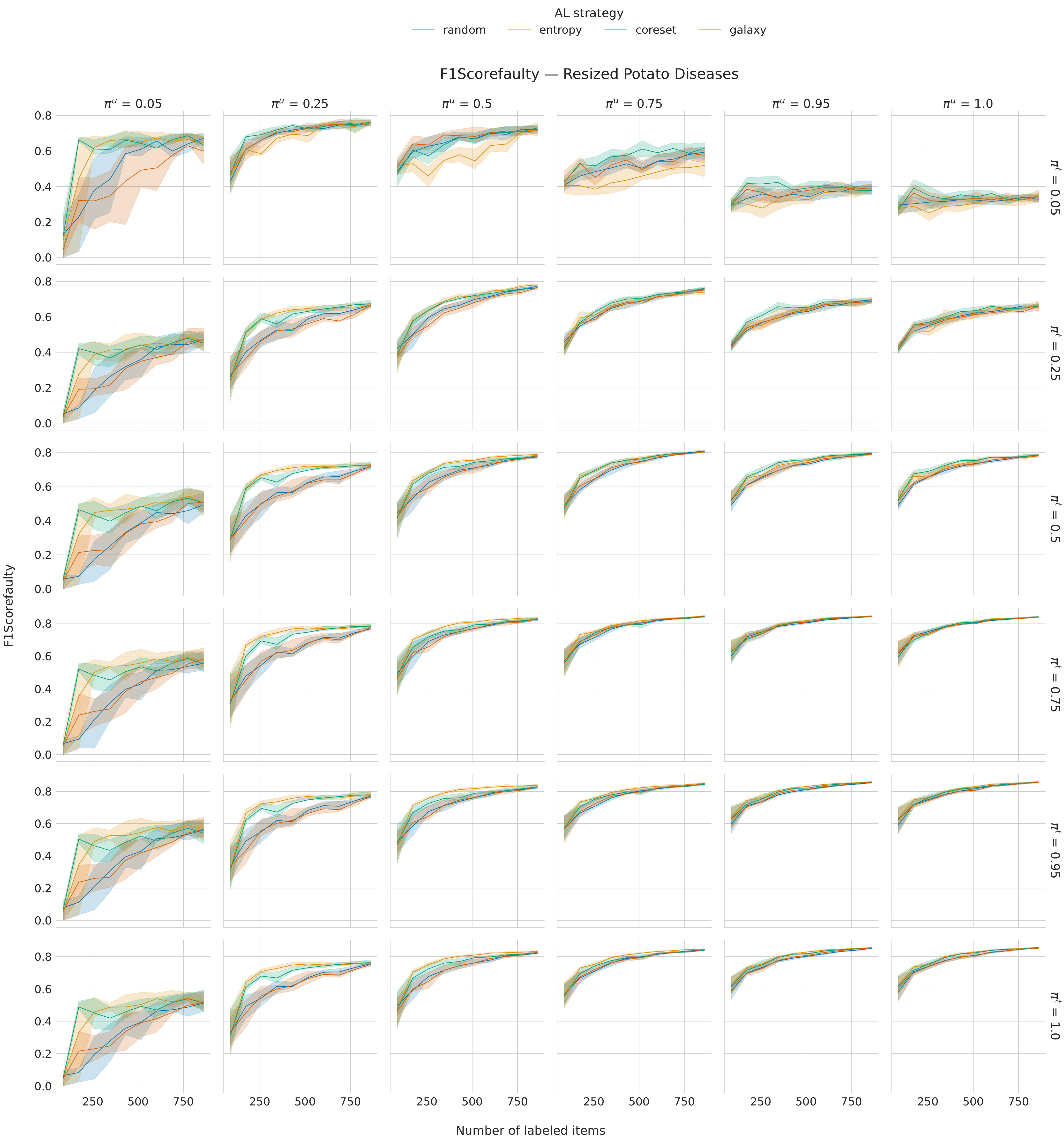}
    \caption{Complete AL results for each pairs of $\pi^u$ | $\pi^t$ measuring F1 Score of the faulty class for the \textbf{resized potato diseases dataset}.
    Unit graphs depict the relation of F1 score for the faulty class along the labelling cycles. The global row axis represent the test set distributions. The column axis represent the distribution of the unlabelled dataset. The diagonal is the classical AL setting where $\pi^u = \pi^t$.}
    \label{fig:complete_resizedPotatof1faulty}
\end{figure}

\begin{figure}[h!]
    \centering
    \includegraphics[height=.8\textheight]{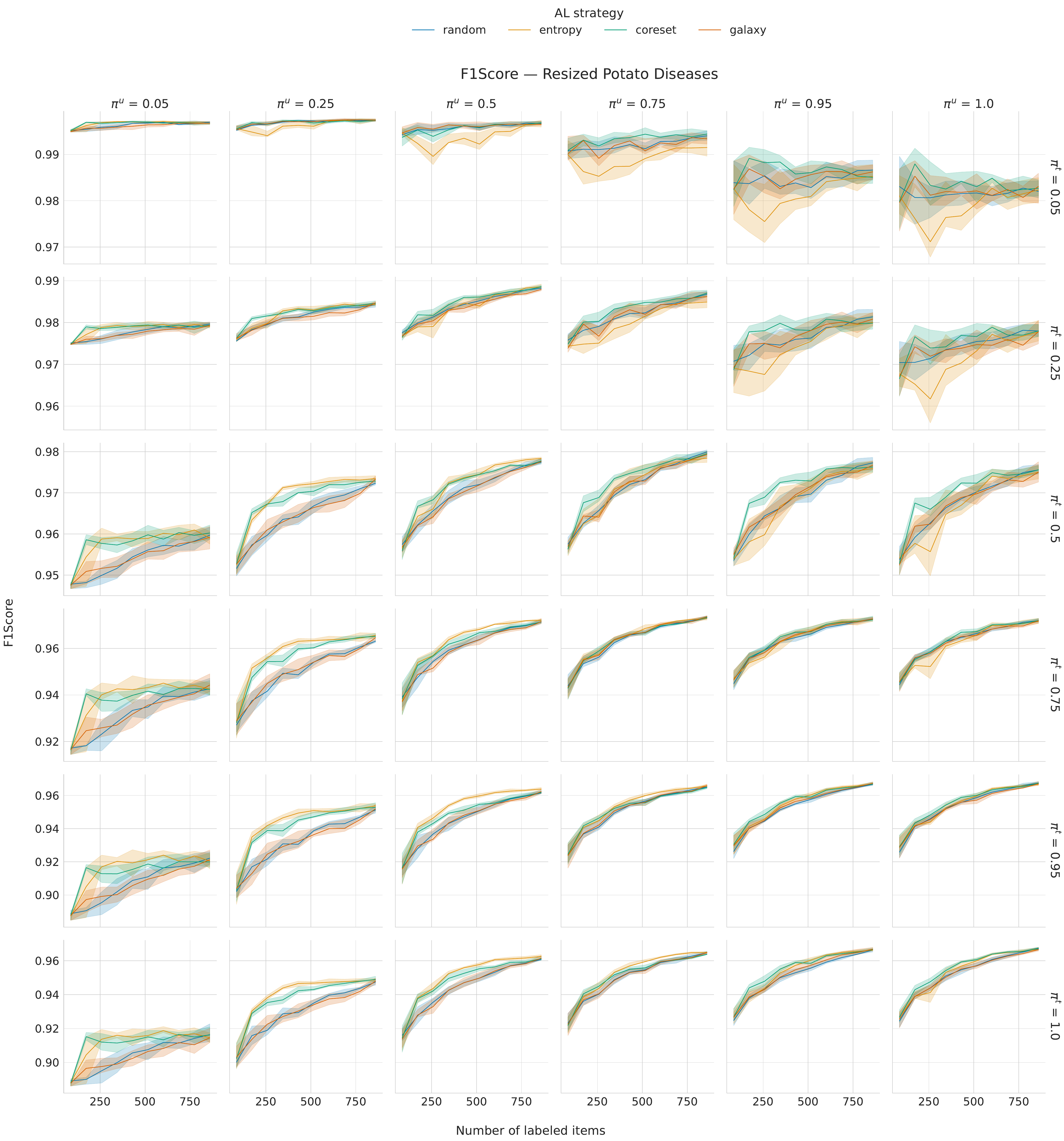}
    \caption{Complete AL results for each pairs of $\pi^u$ | $\pi^t$ measuring Mean IoU for the \textbf{resized potato diseases dataset}.
    Unit graphs depict the relation of F1 score for the faulty class along the labelling cycles. The global row axis represent the test set distributions. The column axis represent the distribution of the unlabelled dataset. The diagonal is the classical AL setting where $\pi^u = \pi^t$.}
    \label{fig:complete_resizedPotatof1}
\end{figure}

\begin{figure}[h!]
    \centering
    \includegraphics[height=.8\textheight]{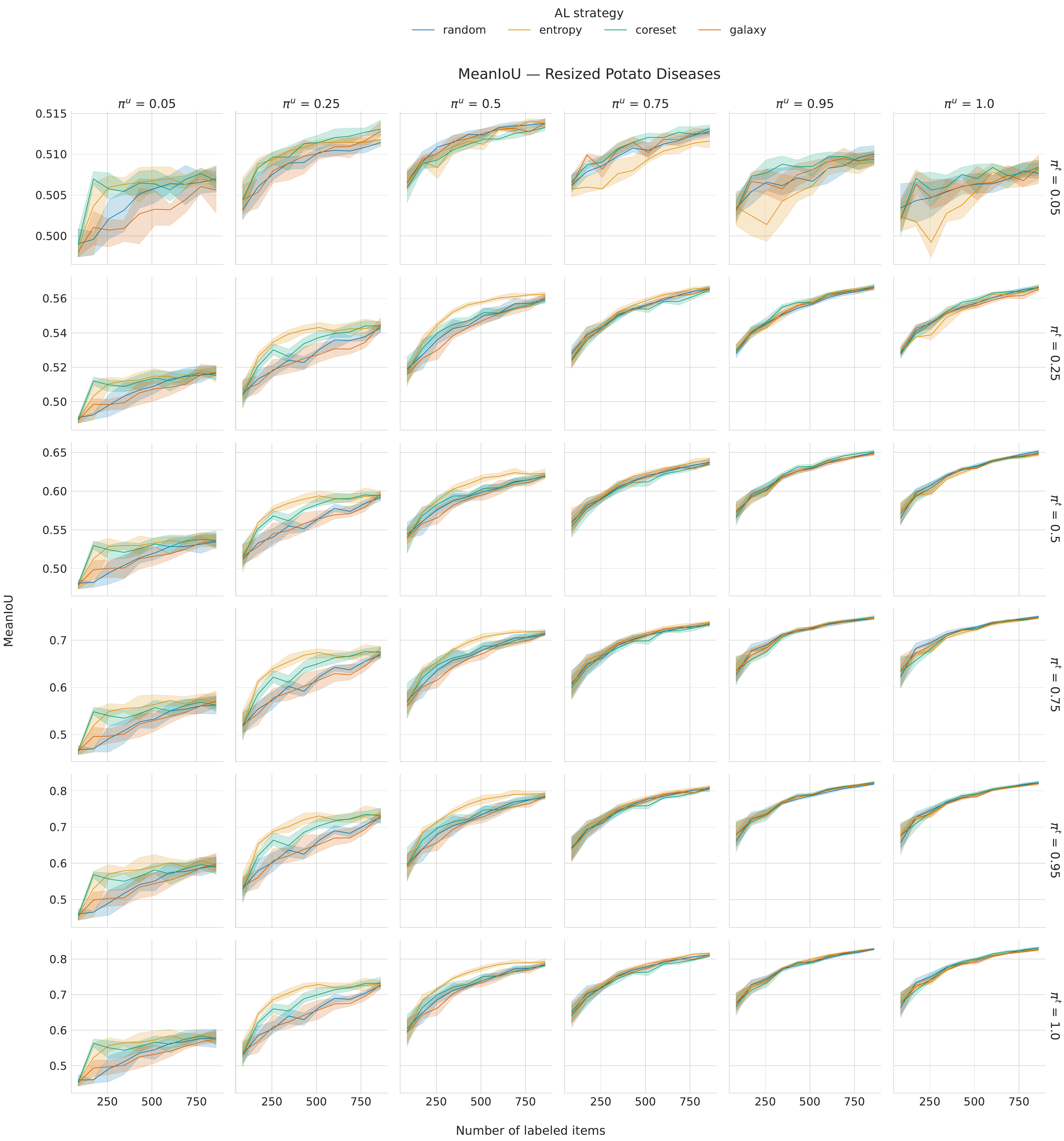}
    \caption{Complete AL results for each pairs of $\pi^u$ | $\pi^t$ measuring Mean IoU for the \textbf{resized potato diseases dataset}.
    Unit graphs depict the relation of F1 score for the faulty class along the labelling cycles. The global row axis represent the test set distributions. The column axis represent the distribution of the unlabelled dataset. The diagonal is the classical AL setting where $\pi^u = \pi^t$.}
    \label{fig:complete_resizedPotatomeanIoU}
\end{figure}

\begin{figure}[h]
\begin{tabular}{|c|}
\hline
\includegraphics[width=\textwidth]{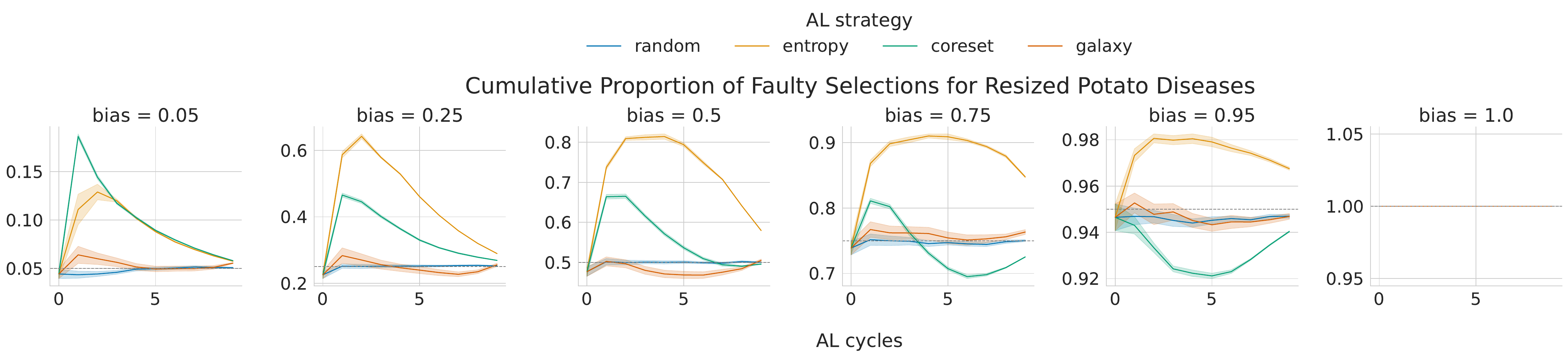}\\
(a) Cumulative proportion of faulty images selected over each AL cycles\\
\hline
\includegraphics[width=\textwidth]{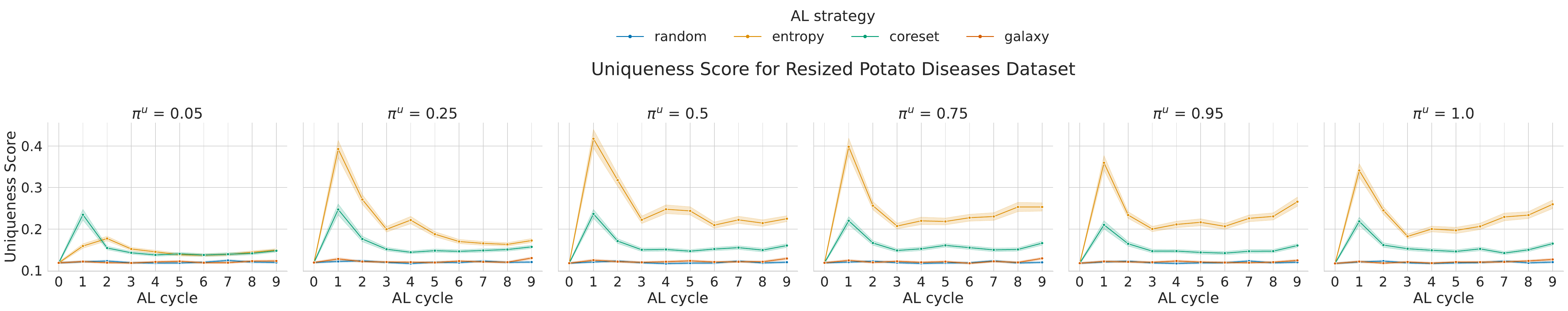}\\
(b) Uniqueness scores \\
\hline
\end{tabular}

\caption{
(a) Proportion of faulty images selected at each cycle for the \textbf{resized potato diseases} dataset. 
(b) Uniqueness scores for the {\bf resized potato diseases} dataset defined by~\eqref{eq:us}. Remind that a uniqueness score of 0.5 means that among all images selected at cycle $j$, half of them are unique.}
\label{fig-defectcum_resizedpotato} 
\end{figure}

\section{Patched Severstal complete results plots}

\begin{figure}[h]
    \centering
    \includegraphics[width=\textwidth]{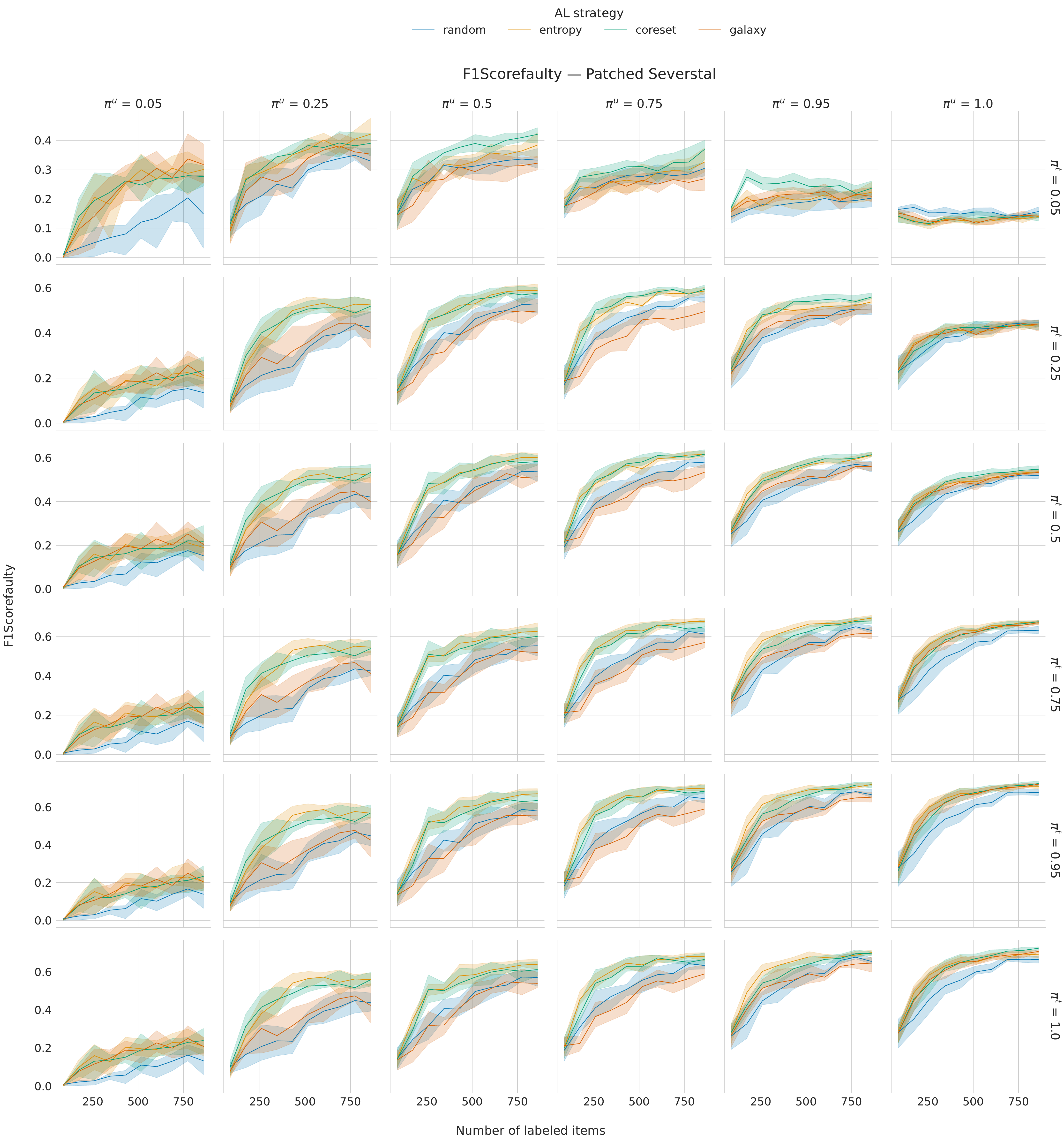}
    \caption{Complete AL results for each pairs of ($\pi^u$, $\pi^t$) measuring F1 Score of the faulty class for the \textbf{patched Severstal} dataset}
    \label{fig-complete_severstal_f1faulty}
\end{figure}

\begin{figure}[h]
    \centering
    \includegraphics[width=\textwidth]{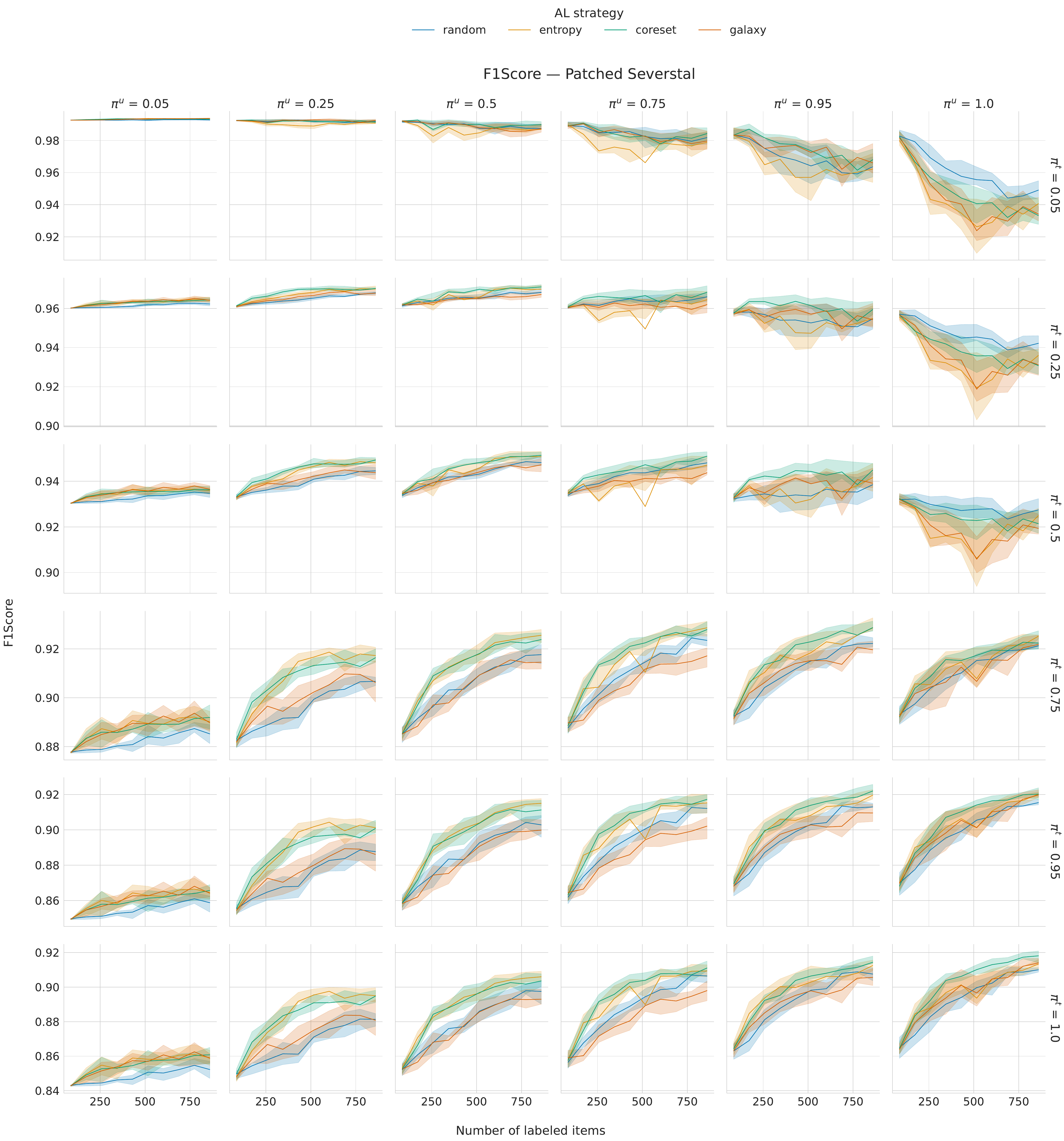}
    \caption{Complete AL results for each pairs of ($\pi^u$, $\pi^t$) measuring average F1 Score for the \textbf{patched Severstal} dataset}
    \label{fig-complete_severstal_f1}
\end{figure}

\begin{figure}[h]
    \centering
    \includegraphics[width=\textwidth]{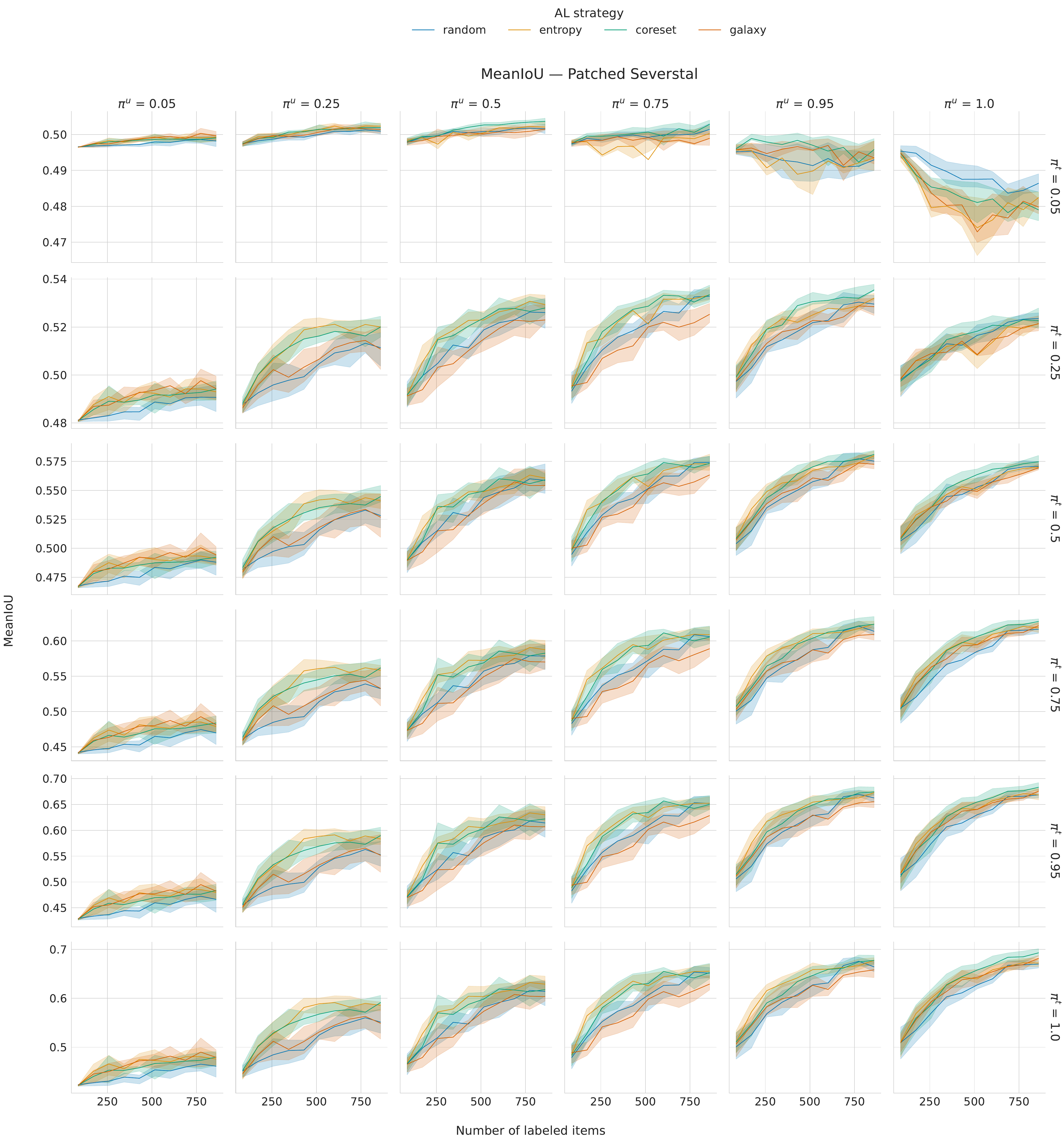}
    \caption{Complete AL results for each pairs of ($\pi^u$, $\pi^t$) measuring MeanIoU for the \textbf{patched Severstal} dataset}
    \label{fig-complete_severstal_meaniou}
\end{figure}

\begin{figure}[h]
\begin{tabular}{|c|}
\hline
\includegraphics[width=\textwidth]{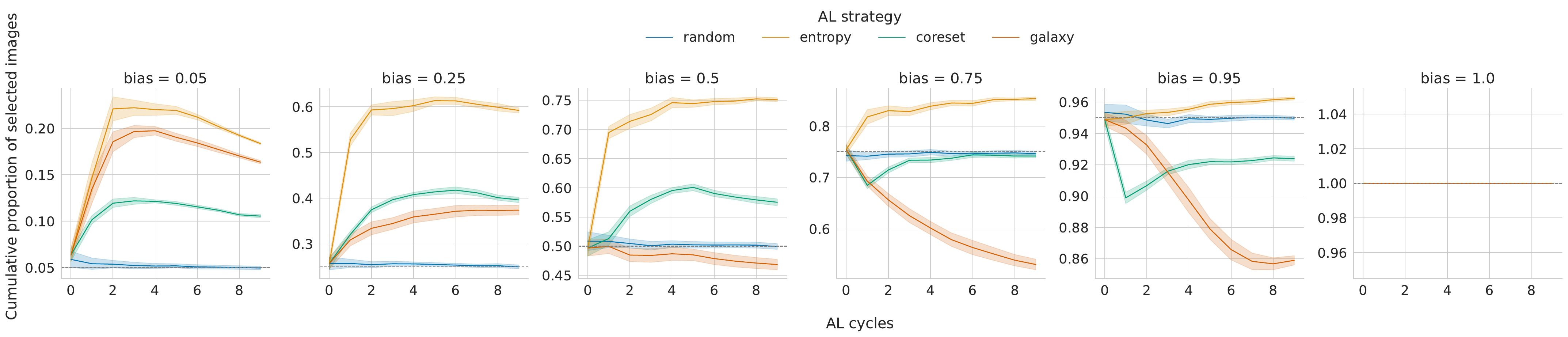}\\
(a) Cumulative proportion of faulty images selected over each AL cycles\\
\hline
\includegraphics[width=\textwidth]{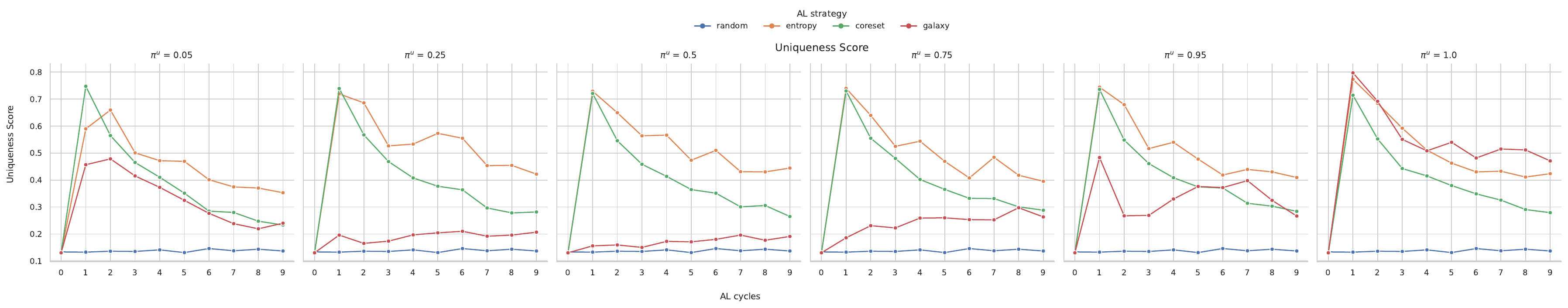}\\
(b) Uniqueness scores \\
\hline
\end{tabular}

\caption{
(a) Proportion of faulty images selected at each cycle for the \textbf{patched Severstal dataset}. 
(b) Uniqueness scores for the {\bf patched Severstal} dataset defined by~\eqref{eq:us}. Remind that a uniqueness score of 0.5 means that among all images selected at cycle $j$, half of them are unique.}
\label{fig-defectcum_severstal} 
\end{figure}

\newpage

\section{Results shared by the three datasets}

\begin{figure}[h]
    \centering
    \includegraphics[width=\textwidth]{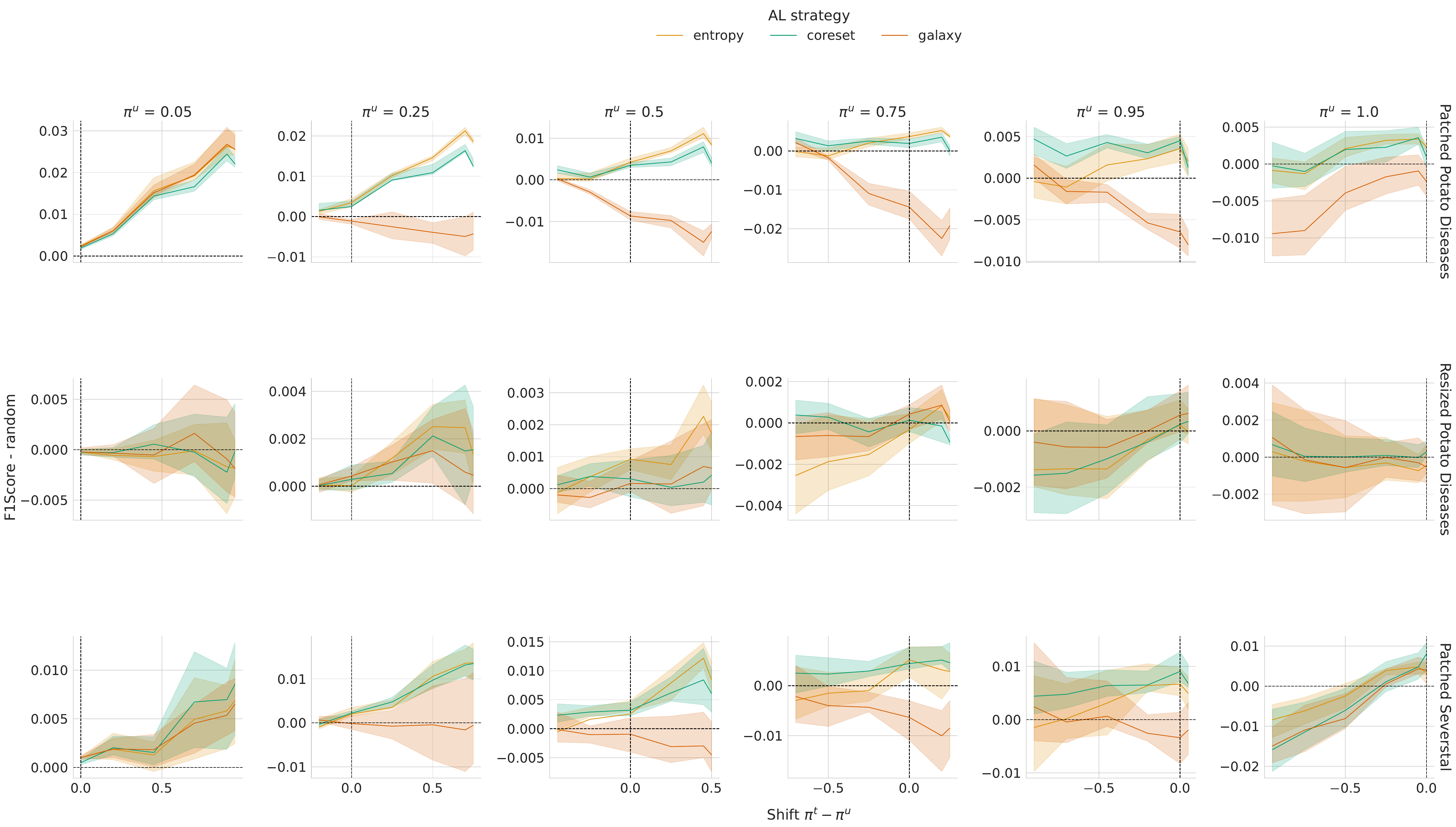}
    \caption{Same as \ref{fig-perf_f1faulty} but measuring average F1 score. Difference of f1 score of each AL strategies (colors) against random sampling, separated for each datasets (rows) and each values of $\pi^u$ (rows). Curves above zero means that AL beats random sampling. x-axis represents the shift between unlabelled pool distribution and test distribution, the lefter the more defect there are in the test set.}
    \label{fig-perff1}
\end{figure}
\begin{figure}[h!]
    \centering
    \includegraphics[width=\textwidth]{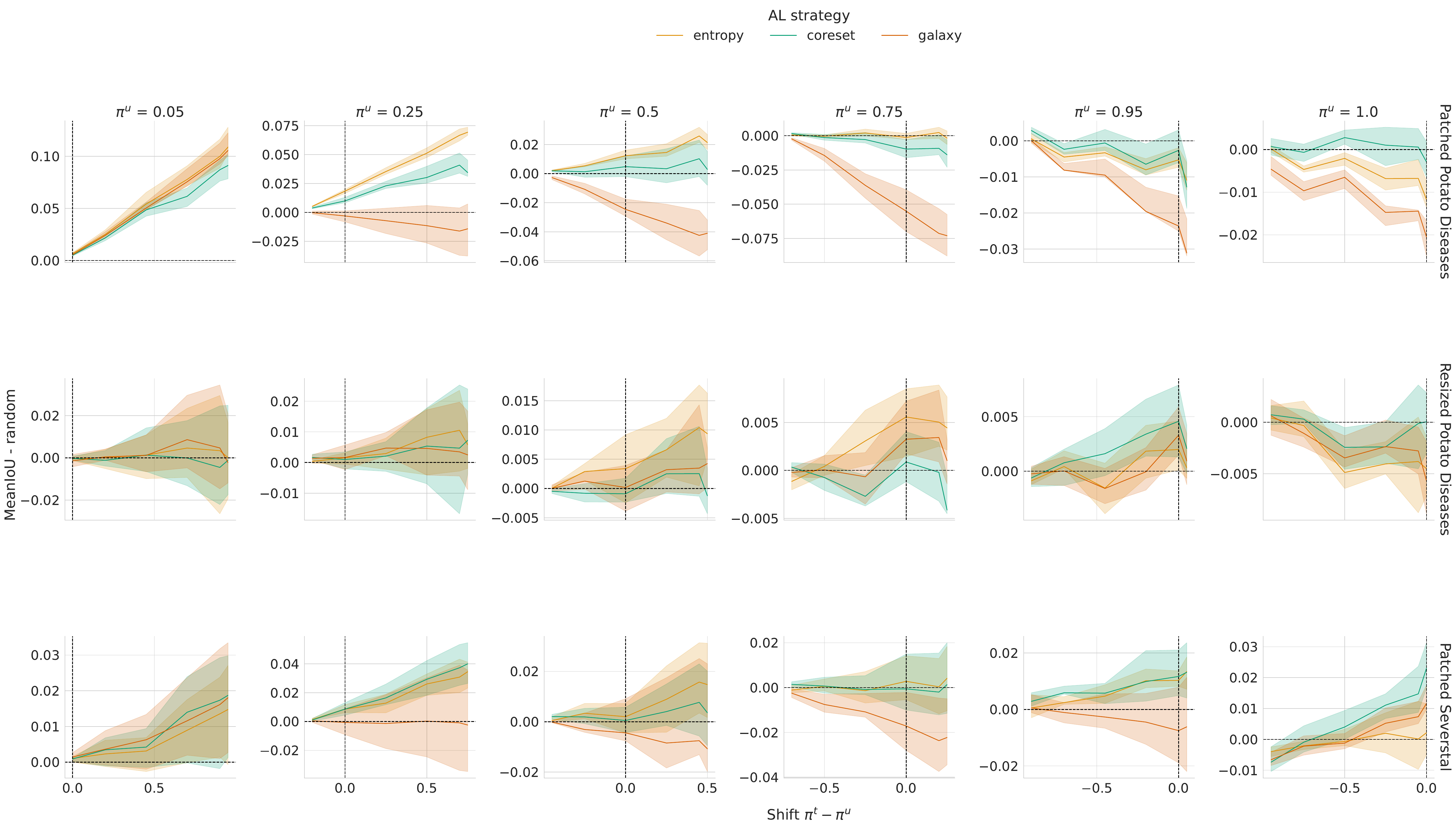}
    \caption{Same as \ref{fig-perf_f1faulty} but measuring MeanIoU. Difference of Mean IoU for the faulty class of each AL strategies (colors) against random sampling, separated for each datasets (rows) and each values of $\pi^u$ (rows). Curves above zero means that AL beats random sampling. x-axis represents the shift between unlabelled pool distribution and test distribution, the lefter the more defect there are in the test set.}
    \label{fig-perfmeaniou}
\end{figure}

\begin{figure}[h]
    \centering
    \includegraphics[width=\textwidth]{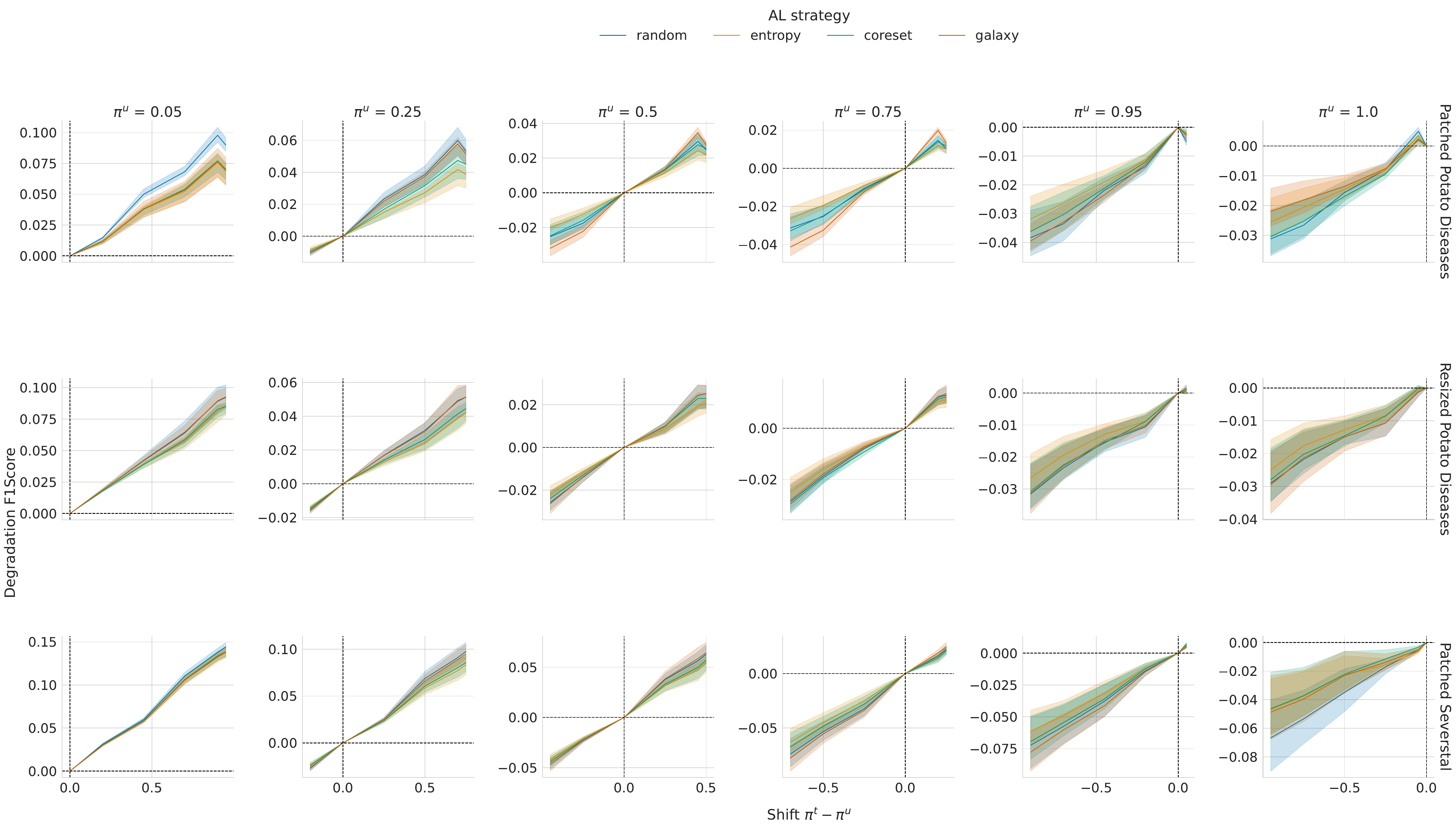}
    \caption{Same as \ref{fig-shift_f1faulty} but measuring average F1 score. Difference of f1 score of each AL strategies (colors) against random sampling, separated for each datasets (rows) and each values of $\pi^u$ (rows). Curves above zero means that AL beats random sampling. x-axis represents the shift between unlabelled pool distribution and test distribution, the lefter the more defect there are in the test set.}
    \label{fig-shiftf1}
\end{figure}

\begin{figure}[h!]
    \centering
    \includegraphics[width=\textwidth]{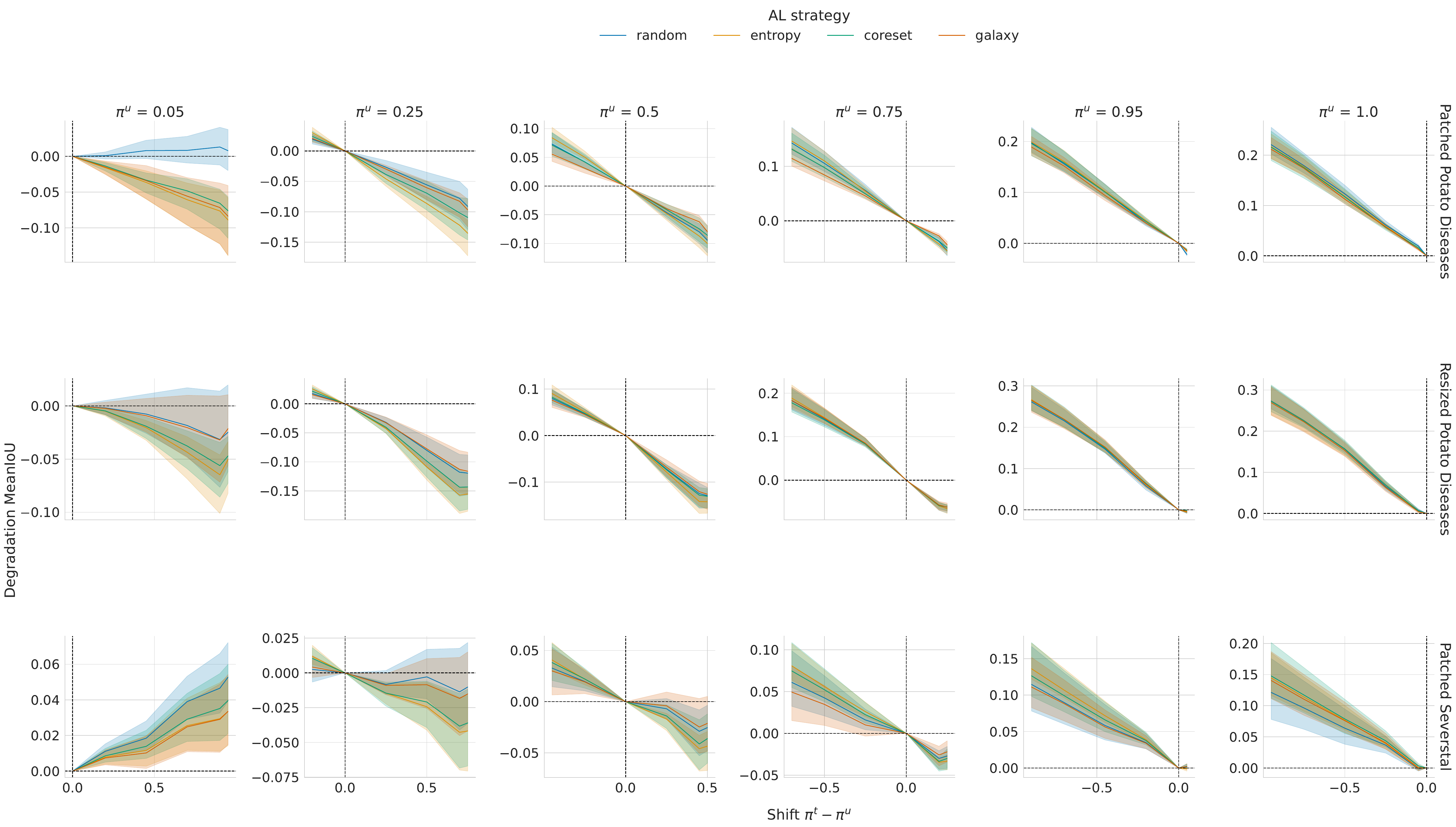}
    \caption{Same as \ref{fig-shift_f1faulty} but measuring MeanIoU. Difference of Mean IoU of each AL strategies (colors) against random sampling, separated for each datasets (rows) and each values of $\pi^u$ (rows). Curves above zero means that AL beats random sampling. x-axis represents the shift between unlabelled pool distribution and test distribution, the lefter the more defect there are in the test set.}
    \label{fig-shiftmeaniou}
\end{figure}

\end{document}